\documentclass[12pt]{article}

\usepackage[margin=1in]{geometry}
\usepackage{amsmath}
\usepackage{amsthm}
\usepackage{amsfonts}
\usepackage{xcolor}
\usepackage{times}
\usepackage{epsfig}
\usepackage{graphicx}
\usepackage{amsmath}
\usepackage{amssymb}
\usepackage[caption=false]{subfig}
\usepackage{bm}
\usepackage[normalem]{ulem}
\usepackage{multirow}
\usepackage{tabu}
\usepackage{enumitem}
\usepackage{adjustbox}
\usepackage{cite}

\def\bbeta{\bm{\beta}}

\def\btheta{\bm{\theta}}
\def\beps{\bm{\epsilon}}

\def\reals{\mathbb{R}}

\def\by{\bm{y}}
\def\bz{\bm{z}}

\def\bw{\bm{w}}

\def\bA{\bm{A}}
\def\bB{\bm{B}}

\def\bX{\bm{X}}

\def\bP{\bm{P}}
\def\bU{\bm{U}}

\def\bD{\bm{D}}
\def\bS{\bm{S}}

\def\bQ{\bm{Q}}

\def\bT{\bm{T}}

\def\cS{\mathcal{S}}

\def\bbR{\bm{R}}
\def\bI{\bm{I}}

\def\T{\top}

\def\ie{{\em i.e.,~}}

\def\bstar{\bbeta^\star}
\def\bhat{\widehat\bbeta}

\def\bt{\tilde{\bbeta}}
\def\grad{\nabla}
\newcommand{\btj}[1]{\bt^{(#1)}}
\newcommand{\bj}[1]{\bbeta^{(#1)}}

\DeclareMathOperator*{\argmin}{arg\,min}

\DeclareMathOperator{\spn}{span}

\newtheorem{myTheorem}{Theorem}
\newtheorem{myLemma}{Lemma}
\newtheorem{myCorollary}{Corollary}

\begin{document}

\title{Neumann Networks for\\ Linear Inverse Problems in Imaging}

\author{
Davis Gilton, Greg Ongie, Rebecca Willett%
\thanks{D.~Gilton is with the Department of Electrical and Computer Engineering, University of Wisconsin, Madison, WI, 53706 USA (e-mail: gilton@wisc.edu). G.~Ongie is with the Department of Statistics, University of Chicago, Chicago, IL, 60637 USA (e-mail: gongie@uchicago.edu). R.~Willett is with the Department of Statistics and Computer Science, University of Chicago, Chicago, IL, 60637 USA (e-mail: willett@uchicago.edu).}
}

\maketitle

\begin{abstract}
Many challenging image processing tasks can be described by an ill-posed linear inverse problem: deblurring, deconvolution, inpainting, compressed sensing, and superresolution all lie in this framework. Traditional inverse problem solvers minimize a cost function consisting of a data-fit term, which measures how well an image matches the observations, and a regularizer, which reflects prior knowledge and promotes images with desirable properties like smoothness. Recent advances in 
machine learning and image processing
have illustrated that it is often possible to \emph{learn} a regularizer from training data that can outperform more traditional regularizers. We present an end-to-end, data-driven method of solving inverse problems inspired by the Neumann series, which we call a Neumann network. 
Rather than unroll an iterative optimization algorithm, we truncate a Neumann series which directly solves the linear inverse problem with a data-driven nonlinear regularizer. The Neumann network architecture outperforms traditional inverse problem solution methods, model-free deep learning approaches, and state-of-the-art unrolled iterative methods on standard datasets. Finally, when the images belong to a union of subspaces  and under appropriate assumptions on the forward model, we prove there exists a Neumann network configuration that well-approximates the optimal oracle estimator for the inverse problem and demonstrate empirically that the trained Neumann network has the form predicted by theory.
\end{abstract}

\section{Learning to Regularize}
In this paper we consider solving linear inverse problems in imaging in which a $p$-pixel image, $\bstar \in \reals^p$  (in vectorized form), is observed via $m$ noisy linear projections as $\by = \bX\bstar + \beps$, where $\by,\beps \in \reals^m$ and $\bX \in \reals^{m\times p}$. This general model is used throughout computational imaging, from basic image restoration tasks like deblurring, super-resolution, and image inpainting \cite{dip}, to a wide variety of tomographic imaging applications, including common types of magnetic resonance imaging \cite{fessler2010model}, X-ray computed tomography \cite{elbakri2002statistical}, radar imaging \cite{blahut2004theory}, among others \cite{barrett2013foundations}. The task of estimating $\bstar$ from $\by$ is often referred to as {\em image reconstruction}. Classical image reconstruction methods assume some prior knowledge about $\bstar$ such as smoothness \cite{tychonoff1977solution}, sparsity in some dictionary or basis \cite{figueiredo2005bound,mairal2009online,yu2012solving}, or other geometric properties \cite{rudin1992nonlinear,willett2003platelets,danielyan2012bm3d,marais2017proximal}, and attempt to estimate a $\bhat$ that is 
both a good fit to the observation $\by$ and that also conforms to this prior knowledge. In general, a {\em regularization function} $r(\bbeta)$ measures the lack of conformity of $\bbeta$ to this prior knowledge and $\bhat$ is selected so that $r(\bhat)$ is as small as possible while still providing a good fit to the data. 

However, recent work in computer vision using deep neural networks has leveraged large collections of ``training'' images to yield unprecedented image recognition performance \cite{he2016deep, huang2017densely, larsson2016fractalnet}, and an emerging body of research is exploring whether this training data can also be used to improve the quality of image reconstruction. In other words, {\em can training data be used to learn how to regularize inverse problems?} As we detail below, existing methods include using training images to learn a low-dimensional image manifold and constraining $\bhat$ to lie on this manifold \cite{bora2017compressed} or learning a denoising autoencoder that can be treated as a regularization step (\ie proximal operator) within an iterative reconstruction scheme \cite{meinhardt2017learning}. 

In this paper, we propose a novel neural network architecture based on the Neumann series expansion \cite{schafheitlin1908theorie,gohberg2013basic} that we call a {\em Neumann network}, describe several of its key theoretical properties, and empirically illustrate its superior performance on a variety of reconstruction tasks. In particular, 
\begin{itemize}[leftmargin=*]
    \item Neumann networks, which directly incorporate the forward operator $\bX$ into the network architecture, can have dramatically lower sample complexity than {\em model-agnostic} networks that attempt to learn the entire image space. As a result, they are much more amenable to applications such as medical imaging or scientific domains where datasets may be smaller.
    \item Neumann networks naturally yield a block-wise structure with {\em skip connections} \cite{he2016deep} emanating from each block. These skip connections appear to yield a smoother optimization landscape that is easier to train than related network architectures.
    \item When the images of interest lie on a union of subspaces, and when the trainable nonlinear components of the network have sufficient expressiveness/capacity, there exists
   a Neumann network estimator that approximates the optimal oracle estimator arbitrarily well. Furthermore, after training the Neumann network on simulated data drawn from a union of subspaces, we show the learned nonlinear components in the trained Neumann network have the form predicted by theory.
    \item A simple preconditioning step combined with the Neumann network further improves empirical performance.
    \item The empirical performance of the Neumann network on superresolution, deblurring, compressed sensing, and inpainting problems exceeds that of competing methods.
\end{itemize}

\section{Previous Work}\label{sec:previous}

There are several general categories of methods used to learn to solve inverse problems, which are reviewed below. Throughout, we assume we have training samples of the form $(\bbeta_i,\by_i)$ for $i = 1,\ldots,N$, where $\by_i = \bX \bbeta_i + \beps_i$, $\bX \in \reals^{m \times p}$ is known and the noise $\beps_i$ is treated as unknown.

\subsection{Agnostic}
An agnostic learner uses the training data to learn a mapping from $\by$  to $\bbeta$ without any knowledge of $\bX$ at any point in the training or testing process \cite{xu2014deep}. 
The general principle is that, given enough training data, we should be able to learn everything we need to know about $\bX$ to successfully estimate $\bbeta$. Empirically,  the success of this approach appears to be highly dependent on the forward operator $\bX$. This straightforward approach has been demonstrated on superresolution \cite{dong2014learning, ledig2017photo}, blind deconvolution \cite{xu2014deep}, and motion deblurring \cite{tan2017motion}, among others. In general, this approach requires large quantities of training data because it is required to not only learn the geometry of the image space containing the $\bbeta$'s, but also aspects of $\bX$. A particularly successful approach to solving inverse problems with neural networks has been through residual learning \cite{he2016deep}.

\subsection{Decoupled}
\label{sec:decoupled}
A decoupled approach operates in two stages. In the first stage, a collection of training images $\bbeta_i$ is used to learn a representation of the image space of interest. In the second stage, this learned representation is incorporated into a mapping from $(\by,\bX)$ to $\bhat$. That is, the learning takes place in a manner that is decoupled from the inverse problem at hand. Here we present two examples of this.

First, we might learn a generative model $G$ for $\bbeta$'s that takes as input a low-dimensional vector $\bz \in \reals^d$ for $d < p$ and outputs $\bbeta = G(\bz)$. The basic idea is that the images of interest lie on a low-dimensional submanifold that can be indexed by $\bz$. Given the learned $G$, we can compute $\bhat$ from $\by$ via

\begin{align}
    \bhat = \argmin_{{\bbeta = G(\bz), \bz \in \reals^d}} \| \by - \bX \bbeta \|_2^2.
\end{align}
This approach was described in \cite{bora2017compressed} with compelling empirical performance for compressed sensing.

Alternatively, we might learn a denoising autoencoder that could be used as a proximal operator in an iterative reconstruction method. Specifically, imagine we had a fixed regularizer $r(\cdot)$ and want to set
\begin{equation}\label{eq:obj}
\bhat = \argmin_{\bbeta} \frac{1}{2}\|\by - \bX \bbeta \|_2^2 + r(\bbeta).
\end{equation}
A proximal gradient algorithm \cite{schmidt2011convergence,parikh2014proximal} starts with an initial estimate $\bbeta^{(0)}$ and step size $\eta > 0$ and then iterates between computing a gradient descent step that pushes the current estimate $\bbeta^{(k)}$ to be a better fit to the data, followed by a {\em proximal operator} that finds an estimate in the proximity of the resulting iterate that is well-regularized (as measured by $r(\cdot)$). This second step is often thought of as a denoising step. One approach to learning to solve inverse problems is to implicitly learn $r(\cdot)$ by explicitly learning a proximal operator in the form of a denoising autoencoder \cite{meinhardt2017learning, chang2017one, gupta2018cnn}. 

The key feature in both of these approaches is that all training takes place independently of $\bX$ -- \ie we either learn a generative model or a proximal operator using the training data, neither of which require knowledge of $\bX$. The advantage of this approach is that once training has taken place, the learned generative model or proximal operator can be used for {\em any} linear inverse problem, so we do not need to re-train a system for each new inverse problem. In other words, the learning is {\em decoupled} from solving the inverse problem.

However, the flexibility of the decoupled approach comes with a high price in terms of sample complexity. To see why, note that learning a generative model or a denoising autoencoder fundamentally amounts to estimating a probability distribution and its support over the space of images; let us denote this distribution as $\phi(\bbeta)$. Thoroughly understanding the space of images of interest is important if our learned regularizer is to be used for linear inverse problems of which we are unaware during training. 

On the other hand, when we know $\bX$ at training time, then we only need to learn the {\em conditional} distribution $\phi(\bbeta | \bX \bbeta)$ or $\phi(\bbeta | \by)$ \cite{efromovich2007conditional}.

For example, imagine an image inpainting scenario in which we only observe a subset of pixels in the image $\bbeta$. Rather than learn the distribution over the space of all possible images, we only need to learn the distribution over the space of missing pixels conditioned on the observed pixels, $\phi(\bbeta | \bX \bbeta)$. Of course, $\phi(\bbeta | \bX \bbeta)$ can be calculated from $\phi(\bbeta)$ and $\bX$ using Bayes' law, but the latter distribution may lie in a much lower-dimensional space, making it easier to learn with limited data. 

It is well-known that the accuracy of any estimate of $\phi(\bbeta)$ has a minimax rate that scales as ${\cal O}(N^{-\frac{\alpha}{2\alpha+p}})$ with $N$ the number of training samples, $p$ the dimension of $\phi(\bbeta)$, and $\alpha$ a smoothness term  \cite{donoho1996density,delyon1996minimax,lafferty2008minimax}. This scaling is quite restrictive, but if the conditional density only depends on a subset of size $p'$ of the original $p$ coordinates, the rate for estimating the conditional density function is ${\cal O}(N^{-\frac{\alpha}{2\alpha+p'}})$ \cite{efromovich2007conditional}.

The key point is that decoupled approaches (implicitly) require learning the full density $\phi(\bbeta)$, whereas a method that incorporates $\bX$ into the learning process has the potential to simply learn the conditional density $\phi(\bbeta | \bX\bbeta)$, which often can be performed accurately with relatively little training data.  This observation is supported by our experimental results, which illustrate that decoupled approaches generally require far more training samples than methods that incorporate knowledge of $\bX$.

\begin{figure*}[ht!]
    \centering
    \includegraphics[width=\textwidth]{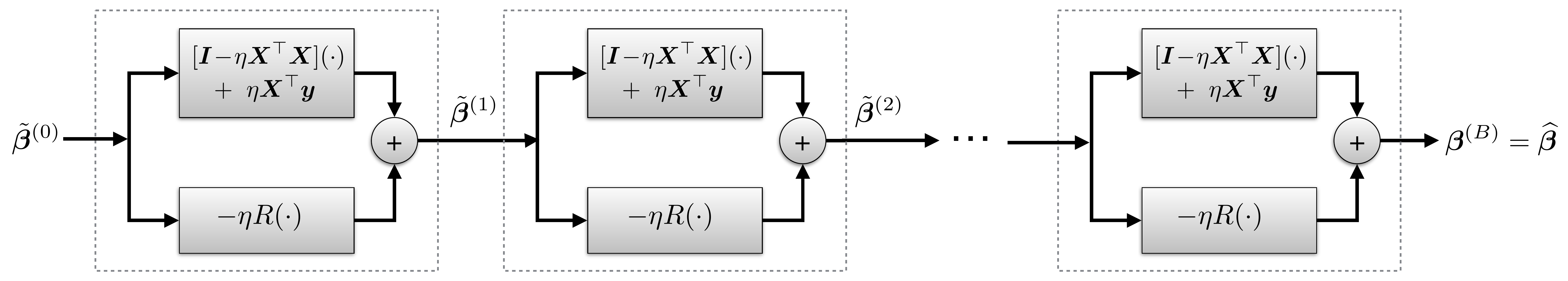}
    \caption{\small Unrolled gradient descent network. The result of $B$ iterations of gradient descent with a fixed step size $\eta$ and regularizer with gradient $R$, as in \eqref{eq:gdn} is equivalent to the output of the above network, with each block corresponding to a single iteration. The network maps a linear function of the measurements, $\bbeta^{(0)} = \eta \bX^\top\by$, to a reconstruction $\bhat$ by successive application of an operator of the form $[\bI-\eta\bX^\T\bX](\cdot)-\eta R(\cdot)$ and addition of $\eta \bX^\top \by$. Here $R$ is a trained neural network, and the scale parameter $\eta$ is also trained.
    }
    \label{fig:GD}
\end{figure*}

\subsection{Unrolled Optimization} \label{sec:unrolled}
Another approach treats a learned component of a network as the gradient of a prior over the data or a proximal operator for a regularizer \cite{chen2017trainable}. Suppose the desired optimal point $\bstar$ satisfies the optimality condition in \eqref{eq:obj}
Now we assume $r(\cdot)$ is differentiable and 
let $R(\bbeta):= \grad r(\bbeta)$ denote the gradient of the regularizer. Then solving \eqref{eq:obj} can be accomplished using iterative optimization; for instance, gradient descent would result in the iterates
\begin{equation}\label{eq:gdn}
\bbeta^{(k+1)} = \bbeta^{(k)} -\eta \left[\bX^\top(\bX \bbeta^{(k)}-\by) + R(\bbeta^{(k)})\right]
\end{equation}
for a step size $\eta > 0$.
Imagine computing these iterates for a fixed number of iterations, which we will denote $B$ (for Blocks, as will become clear shortly). 

``Unrolling" an optimization method refers to taking an iterative optimization method and, instead of iterating until convergence, thinking of a series of $B$ iterates as a single operation to be applied to an input. This idea as applied to gradient descent is represented pictorially  in Figure~\ref{fig:GD}. We can now represent the gradient of the regularizer, $R(\cdot)$, with a trainable neural network. In contrast to the decoupled approach described above, unrolled optimization methods learn the regularizer (or its gradient) in the context of the forward model $\bX$ and training observations $\by_i$ by minimizing the disparity between the true $\bbeta_i$ and $\bhat(\by_i)$, the output of the full network (see Figure~\ref{fig:GD}). This end-to-end training sidesteps the sample complexity challenges described in Section~\ref{sec:decoupled}. 

The unrolling approach can be applied to a variety of optimization algorithms beyond gradient descent. The earliest proposed unrolled inverse problem solver was \cite{Gregor2010LFA}, in which the authors proposed unrolling the Iterative Shrinkage and Thresholding Algorithm (ISTA) \cite{beck2009fast} and the coordinate descent algorithm; 
further refinements of this approach were proposed in \cite{sprechmann2012learning,kamilov2016learning}. More recent work has illustrated the efficacy of unrolled optimization as applied to (proximal) gradient descent \cite{chen2017trainable,diamond2017unrolled,mardani2018neural}, alternating directions method of multipliers \cite{sun2016deep}, primal-dual methods \cite{adler2018learned}, half-quadratic splitting \cite{schmidt2014shrinkage,zhang2017learning}, block coordinate descent \cite{ravishankar2017physics, ravishankar2018deep,chun2018deep}, alternating minimization \cite{aggarwal2018modl}, iterative reweighted least squares \cite{aggarwal2018multi,pramanik2018off}, and approximate message passing \cite{metzler2017learned}. In proximal gradient settings, the learned neural network is interpreted as a learned proximal operator, whereas in the gradient descent network, the learned neural network is interpreted as the gradient of the regularizer at the input. In other words, for different unrolled optimization methods the learned neural network can play different roles.

While for practical reasons the number of blocks $B$ must be kept small in end-to-end training, empirically this does not appear to be an obstacle to good performance. For example, \cite{Gregor2010LFA} notes that end-to-end training reduces the iterations of the ISTA algorithm required to achieve a fixed error rate by a factor of 20, and \cite{diamond2017unrolled} achieve promising performance with $B = 8$ proximal gradient descent iterations. 
Another strategy to enable deeper unrollings is to perform block-by-block training as in \cite{ravishankar2018deep}; however, this approach is unsuitable when the neural network weights are shared between blocks, which is the case in our setting. It is possible to relax the shared-weights assumption, but \cite{aggarwal2018modl} has illustrated that in the low-sample setting, different learned weights in each block can be suboptimal.

\section{Neumann Networks}  
Below, we adopt the following strategy. First, we consider the setting in which the gradient of the regularizer is a linear operator and derive a simple Neumann series approximation to an optimal solution of \eqref{eq:obj}. We then consider the overall {\em Neumann network} formed if $R$ is represented by a (potentially nonlinear) neural network. In this section, we treat a nonlinear network operation as a heuristic that we justify theoretically in Section~\ref{sec:theory}. This section also describes a simple preconditioning step that can improve the accuracy of our approach and an explicit comparison between the proposed Neumann network and the unrolled gradient descent network described in Section~\ref{sec:unrolled}. 

\begin{figure*}[ht!]
\centering
\includegraphics[width=\textwidth]{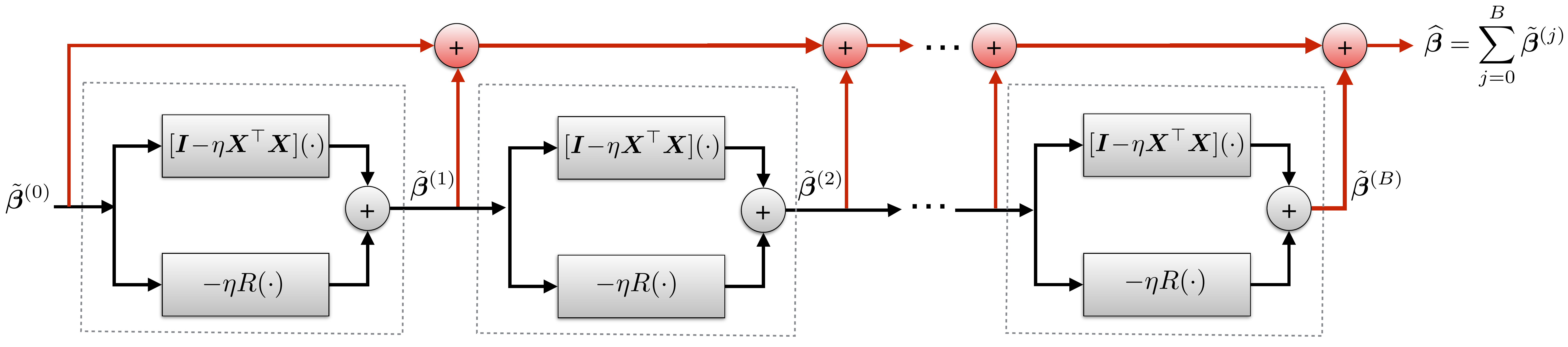}
\caption{\small {Proposed Neumann network  architecture.}  Inspired by the Neumann series expansion for computing the inverse of an operator, a Neumann network maps a linear function of the measurements, $\widetilde \bbeta^{(0)} = \eta \bX^\top\by$ to a reconstruction $\bhat$ by successive application of an operator the form $[\bI-\eta\bX^\T\bX](\cdot)-\eta R(\cdot)$ while summing the intermediate outputs of each block. Here $R$ is a trained neural network, and the scale parameter $\eta$ is also trained. Unlike other networks based on unrolling of iterative optimization algorithms, the series structure of Neumann networks lead naturally to skip connections \cite{he2016deep} (highlighted in red) that route the output of each dashed block to directly to the output layer. }
\label{fig:nn}
\end{figure*}

\subsection{Proposed Network Architecture}
Our proposed network architecture is motivated by the regularized least squares optimization problem \eqref{eq:obj} in the special case where the regularizer $r$ is quadratic. In particular, assume $r(\bbeta) = \frac{1}{2}\bbeta^\T\bbR\bbeta$ so that $\nabla r(\bbeta) = \bbR \bbeta$ for some matrix $\bbR \in \reals^{p\times p}$. Then a necessary condition for $\bstar$ to be a minimizer of \eqref{eq:obj} in this case is
\begin{equation}\label{eq:solution}
    (\bX^\T\bX + \bbR)\bstar = \bX^\T\by.
\end{equation} 

Assuming the matrix on the left-hand side is invertible, the solution is given by
\begin{equation}\label{eq:solution2}
    \bstar = (\bX^\T\bX + \bbR)^{-1}\bX^\T\by.
\end{equation}

In order to approximate the matrix inverse in \eqref{eq:solution2} we consider a Neumann series expansion of linear operators \cite{schafheitlin1908theorie,gohberg2013basic}, which we now recall. Let $\bA$ be any $p\times p$ matrix and let $\bI$ denote the $p\times p$ identity matrix. If the \emph{Neumann series} $\sum_{k=0}^\infty \bA^k$ converges then $\bI-\bA$ is invertible and we have
\begin{equation}\label{eq:neumann1}
    (\bI-\bA)^{-1} = \sum_{k=0}^\infty \bA^k = \bI + \bA + \bA^2 + \bA^3 \cdots
\end{equation}
In particular, a sufficient condition for the convergence of the Neumann series is $\|\bA\| < 1$ where $\|\cdot\|$ is the operator norm. We will make use of an alternative form of the same identity:
\begin{equation}\label{eq:neumann2}
    \bB^{-1} = \eta\sum_{k=0}^\infty(\bI-\eta\bB)^k,
\end{equation}
which is obtained through a change of variables.

Applying the Neumann series expansion \eqref{eq:neumann2} to the matrix inverse appearing in \eqref{eq:solution2}, we have\footnote{The series in \eqref{eq:neumann2} is guaranteed to converge if ${\|\bI-\eta\bB\| < 1}$. Hence, the expansion in \eqref{eq:neumann_expand} is valid provided ${\|\bI-\eta(\bX^\T\bX + \bbR)\| < 1}$, which holds if and only if $\bX^\T\bX + \bbR$ is positive definite and ${\eta < \|\bX^\T\bX + \bbR\|^{-1}}$.}
\begin{equation}\label{eq:neumann_expand}
    \bstar = \sum_{j=0}^\infty(\bI-\eta\bX^\T\bX-\eta \bbR)^j(\eta \bX^\T\by).
\end{equation}
Truncating the series in \eqref{eq:neumann_expand} to $B+1$ terms, and replacing multiplication by the matrix $\bbR$ with a general mapping $R:\reals^p \rightarrow \reals^p$, motivates an estimator $\bhat$ of the form
\begin{equation}\label{eq:nn_estimator}
    \bhat(\by) := \sum_{j=0}^B([\bI-\eta\bX^\T\bX](\cdot)-\eta R(\cdot))^j(\eta \bX^\T\by).
\end{equation}

We turn \eqref{eq:nn_estimator} into a trainable estimator by letting $R = R_{\btheta}$ be a trainable mapping depending on a vector of parameters $\btheta \in \reals^q$ to be learned from training data. Specifically, in this work we assume $R_{\btheta}$ is a neural network, where $\btheta$ is a vectorized set of weights and biases that define the network. We also treat the step-size choice $\eta$ as a trainable parameter. The class of estimators $\bhat(\by) = \bhat(\by;\btheta,\eta)$ specified \eqref{eq:nn_estimator} with trainable network $R = R_{\btheta}$ we call \emph{Neumann networks}.

Observe that Neumann networks are motivated by an application of the Neumann series identity in the case where the gradient of the regularizer is a linear operator (or, equivalently, the regularizer is quadratic). However, for a general regularizer $r$ such that $R = \nabla r$ is nonlinear, the Neumann network estimator $\bhat(\by)$ in \eqref{eq:nn_estimator} may not be a good solution to the optimization problem \eqref{eq:obj}. This is because the Neumann series identity \eqref{eq:neumann2} only holds for linear operators. Despite this fact, we show in the next section that a Neumann network estimator is still mathematically justified under certain model assumptions on the data distribution for which the ideal $R$ is \emph{piecewise linear}. For now, we simply treat the Neumann network estimator as a heuristic motivated by case where $R$ is linear.

To see how \eqref{eq:nn_estimator} can be formulated as a network, observe that the terms in \eqref{eq:nn_estimator} have the following recursive form: let the input to the network be $\tilde{\bbeta}^{(0)} := \eta \bX^\T\by$ and define
\begin{equation}\label{eq:btj}
    \tilde{\bbeta}^{(j)} := (\bI-\eta\bX^\T\bX)\tilde{\bbeta}^{(j-1)}-\eta R(\tilde{\bbeta}^{(j-1)})
\end{equation}
for all $j=1,...,B$. Then we have $\bhat(\by) = \sum_{j=0}^B \btj{j}$.

Figure~\ref{fig:nn} shows a block diagram for implementing the Neumann network using the recursion \eqref{eq:btj}. Each block with a dashed boundary in Figure~\ref{fig:nn} represents an application of the operator $[\bI-\eta\bX^\T\bX](\cdot)-\eta R(\cdot)$. Due to its underlying series structure, the Neumann network has several \emph{skip connections} (highlighted in red) that route the output of each dashed block (\ie the $\btj{j}$'s) to the output layer, similar to those found in residual networks \cite{he2016deep} and related architectures \cite{huang2017densely}. These skip connections are a distinguishing feature of Neumann networks compared to networks derived from unrolled optimization approaches, such as unrolled gradient descent (see Figure~\ref{fig:GD}). We hypothesize these additional skip connections result in a smoother optimization landscape relative to other unrolling approaches, which allows for easier training via stochastic gradient descent. See Section \ref{sec:optlandscape} for empirical evidence and more discussion on this point.

\begin{figure*}[ht!]
\centering
\includegraphics[width=\textwidth]{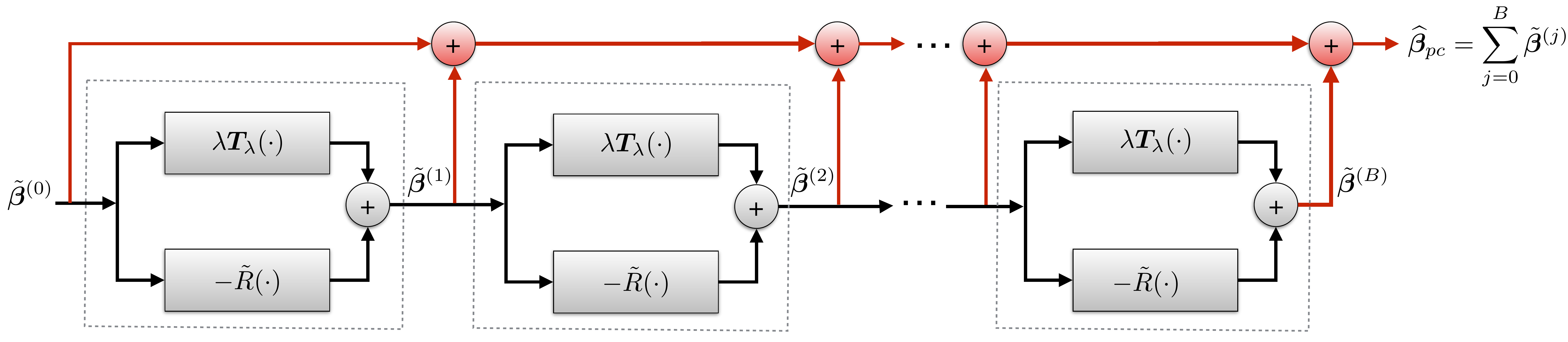}
\caption{\small Proposed preconditioned Neumann network  architecture.  The network has the same basic architecture as a Neumann network, but uses a different linear component given by $\bT_{\lambda} = (\bX^\T\bX + \lambda\bm I)^{-1}$ where $\lambda>0$ and a different initialization $\tilde{\bbeta}^{(0)} = \bT_{\lambda} \bX^\T \by$. When the matrix inverse in $\bT_{\lambda}$ is computationally prohibitive to apply, we replace all instances of $\bT_\lambda$ with an unrolling of a fixed number of iterations of the conjugate gradient algorithm, similar to \cite{aggarwal2018modl}. Here $\tilde{R}$ is a trained neural network, and the scale parameter $\lambda$ is also trained when feasible.}
\label{fig:pnn}
\end{figure*}

\subsection{Preconditioning}
Efficiently finding a solution to the linear system \eqref{eq:solution} using an iterative method is challenging when the matrix $\bX^\T\bX + \bbR$ is ill-conditioned. This suggests that our Neumann network approach, which is derived from a Neumann series expansion of the system in \eqref{eq:solution}, may benefit from preconditioning. Here we derive a variant of Neumann networks inspired by a preconditioning of \eqref{eq:solution}.

Starting from \eqref{eq:solution}, for any $\lambda > 0$ we have
\begin{equation}
    (\bX^\T\bX + \lambda\bI)\bstar + (\bbR-\lambda\bI)\bstar = \bX^\T\by.
\end{equation}
Applying $\bT_\lambda := (\bX^\T\bX + \lambda\bI)^{-1}$ to both sides and rearranging terms gives
\begin{equation}
    (\bI - \lambda\bT_\lambda+\tilde{\bbR})\bstar =  \bT_\lambda\bX^\T\by.
\end{equation}
where we have set $\tilde{\bbR} =  \bT_\lambda \bbR$.
Following the same steps used to derive the Neumann network, we arrive at the modified estimator
\begin{equation}\label{eq:pcnn}
    \bhat_{pc}(\by) = \sum_{j=0}^B(\lambda \bT_\lambda(\cdot)-\tilde{R}(\cdot))^j \bT_\lambda\bX^\T\by
\end{equation}
which we call a \emph{preconditioned Neumann network}. Here $\tilde{R} = \tilde{R}_{\btheta}$ is a trainable mapping depending on parameters $\btheta$. We also treat $\lambda > 0$ as a trainable parameter when gradients with respect to $\lambda$ are easily calculated (more on this below).

As shown in Figure \ref{fig:pnn}, a preconditioned Neumann network has the same basic network architecture as the standard Neumann network, except the linear component ${[
\bI-\eta \bX^\T \bX](\cdot)}$ is replaced with $[\lambda\bT_\lambda](\cdot)$ and the learned component $[-\eta R](\cdot)$ is replaced with $[-\tilde{R}](\cdot)$. The preconditioned Neumann network  also has a different initialization, $\btj{0} =   \bT_\lambda\bX^\T\by$, which is the solution to the Tikhonov regularized least squares problem
    $\min_{\bbeta} \|\bX\bbeta-\by\|^2 + \lambda\|\bbeta\|^2.$
For many inverse problems in imaging, such as deblurring, this is much more accurate approximation to the ideal solution than the matrix transpose initialization $\btj{0} = \eta\bX^\T \by$ of the standard Neumann network. Hence, we might expect that a preconditioned Neumann network  could achieve higher quality solutions with fewer blocks $B$. Our experiments on deblurring of natural images (see Figure~\ref{fig:precondition_bar}) support this observation.

Applying $\bT_\lambda(\cdot)$ may be computationally prohibitive for certain large-scale inverse problems in imaging, such as those arising in CT and MRI reconstruction. To address this issue, we adapt the approach of \cite{aggarwal2018modl} and replace all instances of $\bT_\lambda(\cdot)$ in the preconditioned Neumann network by an unrolling of a fixed number of iterations of the conjugate gradient (CG) algorithm \cite{hestenes1952methods}, which approximates the application of $\bT_\lambda(\cdot)$. Unrolling CG does not require any additional trainable parameters, and backpropagation through the CG layers can be performed via automatic differentiation. This strategy has been shown to be effective for various large-scale MRI reconstruction problems \cite{aggarwal2018multi,pramanik2018off}. Incorporating a trainable $\lambda$ parameter into this approach is simple since the derivatives of the end-to-end network $\bhat_{pc}$ with respect to $\lambda$ are also easily computed by automatic differentiation. In particular, we do not need $\bT_\lambda$ to have an analytic expression in terms of $\lambda$ in order to compute derivatives.

Finally, we note other preconditioned Neumann networks could be derived by replacing $\bI$ with a general matrix $\bS$ such that $\bX^\T\bX + \lambda \bS$ is positive definite, e.g., $\bS = \bD^\T\bD$ where $\bD$ is a discrete approximation of the image gradient. For simplicity, we restrict ourselves to the choice $\bS = \bI$ in this work.

\subsection{Equivalence of Unrolled Gradient Descent and Neumann Network for a Linear Learned Component}\label{sec:equivalence}

Suppose the learned component $R$ is linear, \ie $R(\bbeta) = \bbR\bbeta$ for some matrix $\bbR\in\reals^{p \times p}$.
The $B$th iteration $\bj{B}$ of unrolled gradient descent \eqref{eq:gdn} with step size $\eta>0$ and initialization $\bj{0} = \eta\bX^\T\by$ can be expanded to obtain
\begin{align}\label{eq:equiv}
    \nonumber \bbeta^{(B)} &=  \eta \sum_{j = 0}^{B} (\bI - \eta \bX^\top \bX -\eta \bbR)^j \bX^\top \by,
\end{align}
which is precisely the form of the Neumann network estimator \eqref{eq:nn_estimator}. Therefore, if $R$ is linear the estimator obtained using a unrolling of gradient descent and the Neumann network  estimator are the same. When $R$ is nonlinear we no longer have this equivalence.

\section{Theory}\label{sec:theory}
The Neumann network architecture proposed in the previous section is motivated by the Neumann series expansion of a (potentially nonlinear) operator $R$ representing the gradient of a regularizer. Strictly speaking, this Neumann series expansion is valid (\ie corresponds to the solution of the equation \eqref{eq:obj}) only if $R$ is linear. However, restricting $R$ to to be linear severely limits the class of estimators that can be learned in our framework. In particular, if $R$ is linear then the resulting learned estimator has to be linear, which is suboptimal for many data types.

In this section we show that a Neumann series approach is still mathematically justified for data belonging to a union of subspaces (UoS) model. Our reasons for focusing on a UoS model are two-fold: First, UoS models are a natural generalization of linear subspace models and are used widely in many signal and image reconstruction problems, either as a deterministic model \cite{elhamifar2009sparse,blumensath2011sampling,liu2013robust} or as a statistical model in the form of a Gaussian mixture model with low-rank covariances \cite{renna2014reconstruction,yang2015compressive,houdard2018high}. Second, we believe UoS models represent a reasonable trade-off between model complexity/expressiveness and analytic tractability, and allow us to provide some insight on the expected behavior of Neumann networks beyond the setting where $R$ is linear.

To be precise, here we consider the class of \emph{Neumann network estimators} $\bhat$ given in \eqref{eq:nn_estimator} that are specified by a (potentially nonlinear) mapping\footnote{Here we do not assume $R$ is a neural network with a particular architecture, but study the idealized case where $R$ can represent any mapping from $\reals^p$ to $\reals^p$.} ${R:\reals^p \rightarrow\reals^p}$, step size $\eta$, and number of blocks $B$. We study two questions:
\begin{enumerate}
    \item Can a Neumann network estimator be used to reconstruct images belonging to a UoS, and if so, what is an optimal choice of $R$?
    \item Can this $R$ be learned using standard neural network architectures and training?
\end{enumerate}
Our main result, given in Theorem~\ref{thm:main}, addresses the first question by showing there exists a Neumann network estimator with a \emph{piecewise linear} $R$ that gives arbitrarily small reconstruction error under mild assumptions on the subspaces and their interaction with the measurement operator. We study the second question empirically, and show that the learned $R$ well-approximates the predicted optimal piecewise linear $R$ in an idealized setting.

\subsection{Images Belonging to a Single Subspace}
Suppose the ground truth images belong to an $r$-dimensional subspace $\cS \subset \reals^p$. Let $\bU \in \reals^{p\times r}$ be a matrix whose columns form an orthonormal basis for $\cS$. Assume $m \geq r$ and $\bX\bU \in \reals^{m\times r}$ is full rank. In other words, we assume there is no image in the subspace also in the nullspace of $\bX$ besides the zero image\footnote{This assumption is met in many practical settings. For example, in an inpainting setting it is equivalent to assuming there is no image in the subspace having support contained entirely within the inpainting region. Likewise, in compressed sensing by subsampling DFT coefficients, it is equivalent to assuming there is no image in the subspace bandlimited to the set of unobserved DFT coefficients. Both of these assumptions are reasonable for subspaces spanned by natural images.}. Then given noise-free linear measurements of the form $\by = \bX\bstar$ of any data point $\bstar = \bU\bw^\star \in \mathcal{S}$ we can always recover $\bstar$ by applying the linear estimator
\begin{equation}
    \bhat_o(\by) = \bU(\bU^\T \bX^\T\bX\bU)^{-1}\bU^\T \bX^\T\by
\end{equation}
since it is easy to check that $\bhat_o(\by) = \bstar$. In other words, there always exists a linear estimator that gives exact recovery of images belonging to the subspace from their noise-free linear measurements.

Our first result shows that there exists a linear Neumann network estimator of the form \eqref{eq:nn_estimator} (\ie a linear choice of $R$ in \eqref{eq:nn_estimator}) such that for all points in the subspace the reconstruction error can be made arbitrarily small by choosing the step size $\eta$ and block size $B$ appropriately. For simplicity, we restrict ourselves to the case of noise-free measurements and where $\bX$ has orthonormal rows.

\begin{myLemma}\label{lem:singlesub}
Let $\bX \in \reals^{m\times p}$ be any measurement matrix with orthonormal rows, and let $\cS \subset \reals^p$ be an $r$-dimensional subspace with orthonormal basis $\bU \in \reals^{r\times p}$. Suppose $m\geq r$ and $\bX\bU \in \reals^{m\times r}$ is full rank. Then for any $\eta \in (0,1]$, the $B$-term Neumann network estimator $\bhat$ with linear $R(\bbeta) = \bbR\beta$ where $\bbR \in \reals^{p\times p}$ is given by
\begin{equation}\label{eq:Rform}
\bbR = -c_{\eta,B}\,(\bI-\bX^\T\bX)\bU(\bU^\T\bX^\T\bX\bU)^{-1}\bU^\T\bX^\T\bX
\end{equation}
for a constant $c_{\eta,B}$ depending only on $\eta$ and $B$, 
satisfies the error bounds
\begin{equation}\label{eq:est_bounds}
    \|\bhat(\bX\bstar) - \bstar\| \leq (1-\eta)^{B+1}\|\bX\bstar\|.
\end{equation}
for all $\bstar \in \cS$.
\end{myLemma}

The proof of Lemma 1 is given in the  Appendix. The main idea behind the proof is that with this choice of $R$ the Neumann network terms $\btj{j}$ simplify to
\begin{equation}
    \btj{j} = a_j\bX^\T\bX\bstar + b_j(\bI-\bX^\T\bX)\bstar
\end{equation}
for some constants $a_j$ and $b_j$ that satisfy $\sum_j a_j\approx 1$ and $\sum_j b_j \approx 1$. Hence, we have $\bhat(\by) = \sum_{j=0}^B \btj{j} \approx \bX^\T\bX\bstar + (\bI-\bX^\T\bX)\bstar = \bstar$.

\subsection{Images Belonging to a Union of Subspaces}
Now we suppose that the images belong to a UoS $\cup_{k=1}^K\cS_{k} \subset \reals^p$ where, for simplicity, we assume each subspace $\cS_k$ has dimension $r$. For all $k=1,...,K$ we let $\bU_k \in \reals^{p\times r}$ denote a matrix whose columns form an orthonormal basis for $\cS_k$. Again, we assume $m \geq r$ and $\bX\bU_k \in \reals^{m\times r}$ is full rank for every $k=1,...,K$. In other words, we assume there is no image in the UoS also in the nullspace of $\bX$ besides the zero image.

Let $\by = \bX\bstar$ be the measurements of any point $\bstar$ belonging to the UoS. If we know $\bstar$ belongs to the $k$th subspace, \ie $\bstar = \bU_k\bw^\star$ for some $\bw^\star\in\reals^p$, then similar to the single subspace case, we can apply the estimator
\begin{equation}
    \bhat_o(\by;k) := \bU_k(\bU_k^\T \bX^\T\bX\bU_k)^{-1}\bU_k^\T \bX^\T\by
\end{equation}
since it is easy to see that $\bstar = \bhat_o(\by; k)$. We call $\bhat_o(\by;k)$ the \emph{oracle estimator}, since it assumes knowledge of the subspace index $k$ to which the image belongs. 

We show that, under appropriate conditions on the subspaces and the measurement operator, there is a piecewise linear choice of Neumann network estimator (\ie an estimator of the form \eqref{eq:nn_estimator} with $R$ piecewise linear) that recovers any image belonging to the UoS from its noise-free measurements with arbitrarily small reconstruction error. In other words, there is a Neumann network estimator that well-approximates the oracle estimator.

Specifically, we consider a piecewise linear function $R^*$ of the form
\begin{equation}\label{eq:pwl}
    R^*(\bbeta) = 
    \begin{cases}
    \bbR_1\bbeta & \text{if}~\bbeta\in \mathcal{C}_1\\
    ~~~\vdots & ~~~~ \vdots \\
    \bbR_K\bbeta & \text{if}~\bbeta\in \mathcal{C}_K
    \end{cases}
\end{equation}
where each $\bbR_k$ is a $p \times p$ matrix, and the regions $\mathcal{C}_k$ are disjoint and whose union is all of $\reals^p$. The main idea behind our analysis is this: If the ground truth point $\bstar$ belongs the $k$th subspace, then we prove that the Neumann series summands $\btj{0},\btj{1},...,\btj{B}$ all lie in the same region $\mathcal{C}_k$. This means that the same $\bbR_k$ is used in computing each summand, so we can write
\begin{equation*}
    \bhat(\by) = \sum_{j=0}^B\btj{j} =  \sum_{j=0}^B(\bI - \eta\bX^\T\bX - \eta\bbR_k)^j(\eta\bX^\T\by).
\end{equation*}
Choosing $\bbR_k$ to have the same form as in the single subspace case (see Lemma~\ref{lem:singlesub}), we then will have $\bstar = \bhat_o(\by,k) \approx \bhat(\by)$.

To be exact, we specify $\bbR_k$ and $\mathcal{C}_k$ as follows. Similar to Lemma~\ref{lem:singlesub}, we choose
\begin{equation*}
    \bbR_k \!=\! -c_{\eta,B}\,(\bI-\bX^\T\bX)\bU_{k}(\bU_{k}^\T\bX^\T\bX\bU_{k})^{-1}\bU_{k}^\T\bX^\T\bX,
\end{equation*}
for all $k=1,...,K$, where $c_{\eta,B}>0$ is a constant depending only on $\eta$ and $B$. We also define the corresponding region $\mathcal{C}_k$ as
\begin{equation*}
    \mathcal{C}_k = \left\{\bbeta\in\reals^p: d_{\bX,k}(\bbeta) < d_{\bX,\ell}(\bbeta)~\text{for all}~\ell \neq k \right\}
\end{equation*}
where $d_{\bX,k}(\bbeta) := \|(\bI-\bX\bU_k(\bX\bU_k)^{+})\bX\bbeta\|$ is the distance between the vector $\bX\bbeta$ and the subspace $\spn(\bX\bU_k)$. In other words, $\mathcal{C}_k$ is the set of all points whose distance to the $k$th subspace is smaller than the distance to all other subspaces, as measured by the functions $d_{\bX,\ell}$ for all $\ell =1,...,K$. 

We now state our main theorem:

\begin{myTheorem}\label{thm:main}
Let $\bX \in \reals^{m\times p}$ be any measurement matrix with orthonormal rows, and for all $k=1,...,K$ let $\bU_k \in \reals^{p\times r}$ be an orthonormal basis for the $k$th subspace $\cS_k$ with $\dim \spn(\bX\bU_k) = r$. Suppose $\text{span}(\bX\bU_{k}) \cap \text{span}(\bX\bU_\ell) = \{0\}$ for all $k\neq \ell$. Then the Neumann network estimator $\bhat$ with step size $\eta \in (0,1)$ and piecewise linear $R = R^*$ as defined in \eqref{eq:pwl} satisfies
\begin{equation}\label{eq:est_bounds2}
    \|\bhat(\bX\bstar) - \bstar\| \leq (1-\eta)^{B+1}\|\bX\bstar\|,
\end{equation}
for all $\bstar \in \cup_{k=1}^K\cS_k$.
\end{myTheorem}

The condition $\text{span}(\bX\bU_{k}) \cap \text{span}(\bX\bU_\ell) = \{0\}$ for all $\ell \neq k$, appearing in Theorem~\ref{thm:main} is not overly restrictive if we take into account the statistics of natural images. For instance, this condition holds for a generic union of $r$-dimensional subspaces provided $m \geq 2r$, regardless of the number of subspaces in the union\footnote{This is because if $\spn(\bU_k)$, $k=1,...,K$, are generic $r$-dimensional subspaces in $\reals^p$, then $\mathcal{V}_k = \spn(\bX\bU_k)$, $k=1,...,K$, are generic $r$-dimensional subspaces in $\reals^m$. Since two generic subspaces are linearly independent provided the sum of their dimensions does not exceed the ambient dimension, we see that $\mathcal{V}_k$ and $\mathcal{V}_\ell$, $k\neq \ell$ collectively span a $2r$-dimensional subspace, which is only possible if their intersection is trivial.}. Moreover, based on results in compressive sensing using low-rank Gaussian mixture models \cite{reboredo2013compressive,renna2014reconstruction}, we conjecture this condition can be weakened under appropriate assumptions on $\bX$ and appropriate modification of $R^*$, but we do not pursue this refinement here.

Theorem~\ref{thm:main} shows there exists a Neumann network estimator with a certain choice of $R^*$ that well-approximates an oracle estimator for images belonging to a union of subspaces. In principle, since $R^*$ is piecewise linear with a finite number of regions, it is realizable as a sufficiently deep neural network with ReLU activations \cite{arora2018understanding}. However, this does not necessarily mean that a Neumann network estimator with $R$ given by a ReLU network when trained on images belonging to a union of subspaces will recover $R = R^*$ specified in Theorem 1. For example, there may be other $R'$ that yield similar training loss as $R^*$, or the learned component may be under-parameterized (e.g., not enough layers) in such a way that it cannot well-approximate $R^*$. Nevertheless, one would hope that given sufficient training data and a sufficiently expressive network architecture for the learned component, it may be possible to learn a good approximation to $R^*$ as specified in Theorem~\ref{thm:main}. Below we illustrate that this is indeed the case for a Neumann network trained on images belonging to synthetic union of subspaces.

Finally, using the equivalence of Neumann networks and unrolled gradient descent networks estimators in the case where the learned component $R$ is linear (see Sec. \ref{sec:equivalence}), we show that an unrolled gradient descent network as defined in \eqref{eq:gdn} with the same piecewise linear $R=R^*$ as defined in \eqref{eq:pwl} satisfies the error bounds as in Theorem~\ref{thm:main}:
\begin{myCorollary}\label{cor:main}
Under the same assumptions as Theorem~\ref{thm:main}, the unrolled gradient descent estimator $\bhat'(\by) = \bj{B}$ with step size $\eta \in (0,1)$ and $R = R^*$ as defined in \eqref{eq:pwl} satisfies 
\begin{equation}\label{eq:est_bounds3}
    \|\bhat'(\bX\bstar) - \bstar\| \leq (1-\eta)^{B+1}\|\bX\bstar\|,
\end{equation}
for all $\bstar \in \cup_{k=1}^K\cS_k$,
\end{myCorollary}

Corollary 1 shows that the equivalence between unrolled gradient descent estimators and Neumann network estimators observed in the case where $R$ is linear carries over to the special case where $R = R^*$ is piecewise linear and the networks are evaluated on linear measurements of points belonging to the union of subspaces.

\subsection{Empirical Validation}

Here we illustrate empirically that the optimal $R^*$ predicted by Theorem 1 is well-approximated by training a Neumann network for a 1-D inpainting task on synthetic UoS data. We generate random training data belonging to a union of three  3-dimensional subspaces in $\reals^{10}$, and train a Neumann network to inpaint five missing coordinates (\ie $\bX \in \reals^{5\times 10}$ restricts a vector to coordinates 1--5). We parameterize the learned component $R$ of the Neumann Network as a $7$-layer fully connected neural network with ReLU activations, which is trained by minimizing the mean squared error of the reconstruction over the training set using stochastic gradient descent (more details on this experiment can be found in the Supplementary Materials).

\begin{figure}[ht!]
\centering
\includegraphics[width=0.45\textwidth]{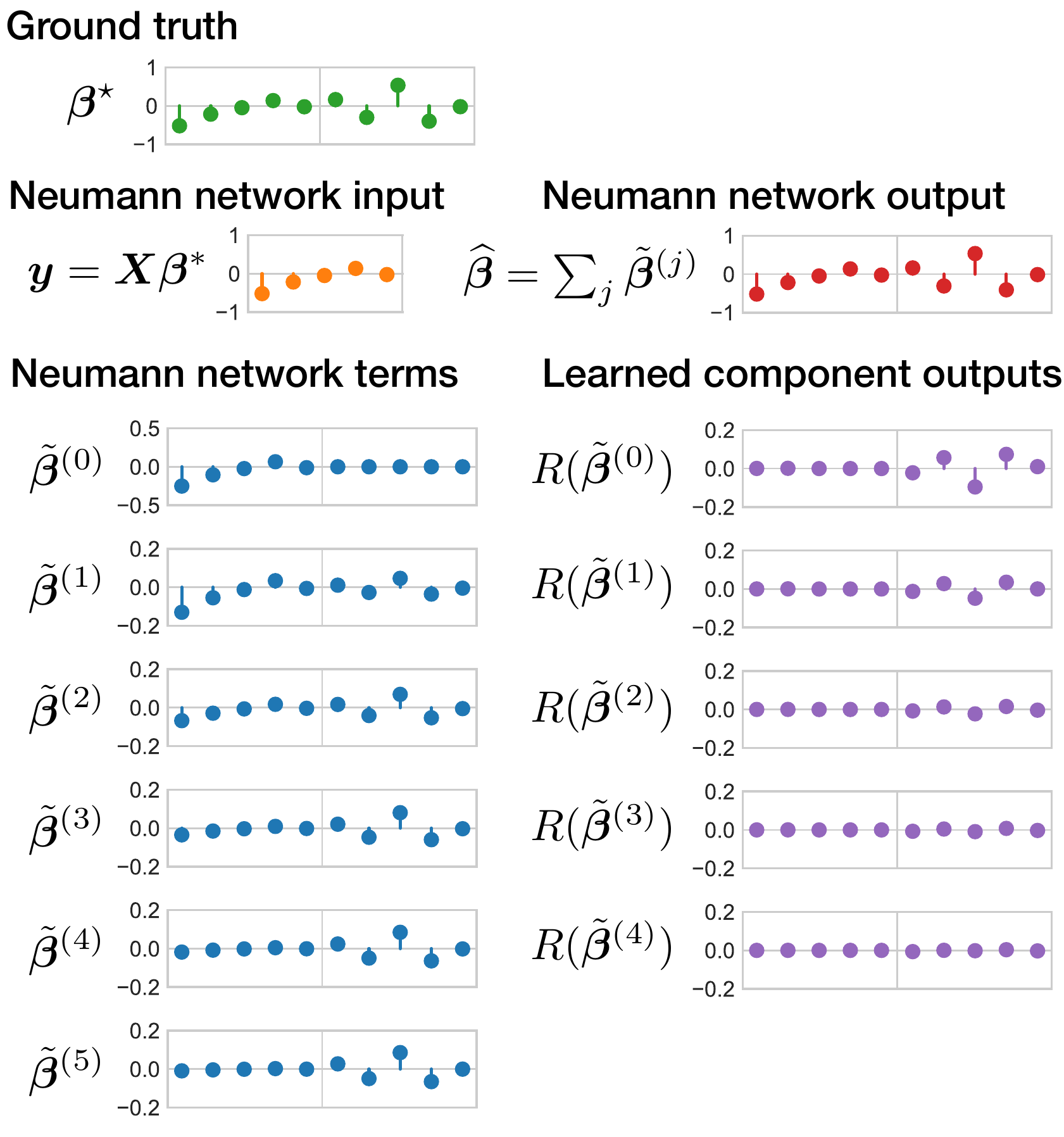}
\caption{\small Example output of Neumann network trained on synthetic union of subspaces data for a 1-D inpainting task. Here a vector $\bstar \in \reals^{10}$ is drawn from one of the subspaces, and its measurements $\by = \bX\bbeta^*$ (restriction to first five coordinates) are input into the Neumann network, which faithfully restores the missing coordinates. The output $\bhat$ of the Neumann network  is a sum of terms $\btj{j}$ (shown in bottom left). 
As predicted by Theorem~\ref{thm:main}, the terms $\btj{j}$ are weighted linear combinations the projections of $\bstar$ onto the observed and unobserved coordinates. Also as predicted by Theorem~\ref{thm:main}, the outputs of the learned component $R(\btj{j})$ (shown in bottom right) are zero in the observed coordinates and scaled projections of $\bstar$ in the unobserved coordinates.} 
\label{fig:Bexp}
\end{figure}

\begin{figure}[ht!]
\centering
\includegraphics[height=0.32\textwidth]{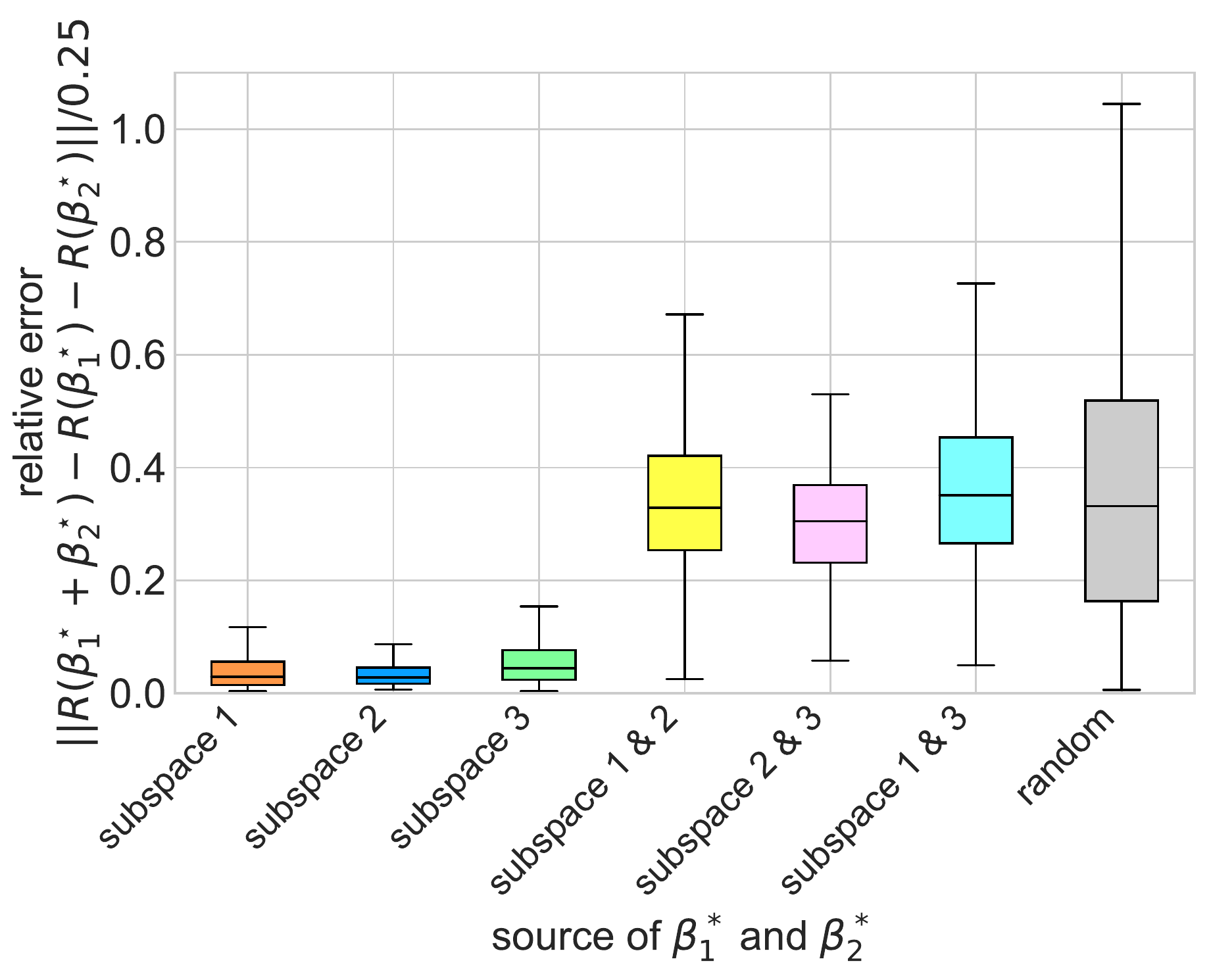}
\caption{\small Piecewise linearity test. We measure how linear the learned $R$ is when evaluated at two vectors drawn from the same subspace, from two different subspaces, or from two random Gaussian vectors. The plot illustrates that the learned $R$ only behaves like a linear operator when the vectors belong the same subspace (\ie the relative error is small), which indicates the learned $R$ is approximately piecewise linear, as predicted by Theorem~1.}
\label{fig:learnedR_exp1}
\end{figure}

Figure~\ref{fig:Bexp} illustrates the output of the trained Neumann network for one specific input, including the outputs from the intermediate Neumann network terms $\btj{j}$ and the learned component outputs $R(\btj{j})$. As predicted by Theorem~\ref{thm:main}, the Neumann network terms $\btj{j}$ have the form $a_j \bX^\T\bX\bstar + b_j (\bI-\bX^\T\bX)\bstar$ for some constants $a_j$ and $b_j$. Also, the outputs of the learned component $R(\btj{j})$ all lie in the null space of $\bX$, \ie are vectors supported on coordinates $6-10$.

Figure~\ref{fig:learnedR_exp1} displays the results of a quantitative experiment to assess whether the learned component $R$ is piecewise linear as predicted by Theorem 1. First, we test whether the learned $R$ is approximately linear when restricted to inputs belonging to each subspace, \ie we test whether $R(\bstar_1 + \bstar_2) \approx R(\bstar_1) + R(\bstar_2)$, for all $\bstar_1,\bstar_2$ belonging to the same subspace. As baselines we compare to the case where $\bstar_1$ and $\bstar_2$ belong to different subspaces, and the case where $\bstar_1$ and $\bstar_2$ are Gaussian random vectors. In Figure~\ref{fig:learnedR_exp1} we display a boxplot of the relative error $\|R(\bstar_1 + \bstar_2) -R(\bstar_1)-R(\bstar_2)\|/\gamma$ of 1024 randomly generated $\bstar_1,\bstar_2$, which are normalized such that $\|\bstar_1\|=\|\bstar_2\| = \gamma$. Here we set normalization to $\gamma = 0.25$, though similar results were obtained for $\gamma \in [0.1,0.5]$ (not shown). As predicted, the relative error concentrates near zero in the case where $\bstar_1,\bstar_2$ belong to the same subspace, and is otherwise large, indicating the learned $R$ is indeed approximately piecewise linear as predicted by Theorem 1.

In the Supplementary Materials we provide more empirical evidence that the $R$ learned in this experiment closely approximates the ideal $R^*$ predicted by Theorem 1. Specifically, we demonstrate that the learned component behaves as expected on inputs restricted to the column space and row space of the forward model $\bX$. These experiments verify that, at least for this 1-D inpainting task on synthetic data, the ideal piecewise linear $R^*$ predicted by Theorem 1 is well-approximated with standard neural network architectures and training.

A more systematic study involving different forward models $\bX$ and different network architectures is needed to determine whether the ideal $R^*$ identified in Theorem~1 is learnable more generally for large-scale imaging data and using practical architectures like convolutional neural networks. Also, our results do not address the case of noisy measurements or forward models with non-orthogonal rows, which are important considerations for many inverse problems. We leave these as open questions for future work.

Finally, while our focus in this section was on UoS models, our analysis does not rule out the applicability of Neumann networks to other non-linear models. Indeed, in the next section we show empirically that Neumann networks perform well on a variety of linear inverse problems when trained on realistic image datasets that are unlikely to be perfectly captured by a low-dimensional UoS model.

\section{Experiments}

We begin this section with a comparison of Neumann networks against other methods of solving several different inverse problems with learned components. 
After that, we investigate the effect of larger and smaller training sets on all methods, demonstrating that Neumann networks are robust to small training set sizes.  
We follow these experiments with an illustration of the effects of incorporating preconditioning into the Neumann network for deblurring, which is shown to give a gain of several dBs of PSNR, permitting smaller networks and allowing for faster training and implementation. We follow with an investigation into an MRI reconstruction problem to demonstrate the proposed methods in a large-scale setting. Finally, we explore the optimization landscape of Neumann networks relative to unrolled gradient descent, illustrating that Neumann networks have smoother loss landscapes than unrolled Gradient Descent, while also generally achieving lower test set errors.

\subsection{Datasets and Comparison Methods}
In our experiments, we consider three different small-scale training sets: CIFAR10 \cite{krizhevsky2009learning}, CelebA \cite{liu2015faceattributes}, and STL10 \cite{coates2011analysis}, and one larger-scale undersampled MRI reconstruction task.

The CIFAR10 dataset is a machine learning standard, consisting of real-world images of both man-made and natural scenes \cite{krizhevsky2009learning}. The dataset has been resized to be $32 \times 32$ pixels.
  
We use a subset of the aligned Celebrity Faces With Attributes (CelebA) dataset \cite{liu2015faceattributes}. The CelebA dataset consists of human faces at a variety of angles, and the subset that is used here has been aligned so that all faces lie in the center of the image.

The STL10 dataset \cite{coates2011analysis} is a curated subset of the ImageNet dataset, and was originally intended to be used with semisupervised learning problems. In our experiments, we have resized all CelebaA and STL10 images to be $64 \times 64$ pixels.

We select a subset of images of size 30,000 uniformly from each individual dataset to be used for training in the results presented below.

We use these training sets in seven different inverse problems in imaging: Block inpainting, deblurring, deblurring with additive noise $\epsilon$ of variance $0.01$, superresolution (SR4 and SR10) with two different upsampling levels (4x and 10x across the entire image, respectively), and compressive sensing (CS2 and CS8) with two separate levels of compression (2x and 8x, respectively). The compressed sensing design matrices are random Gaussian matrices.

We compare Neumann networks \textbf{(NN)} and preconditioned Neumann networks \textbf{(PNN)} with four methods which can be applied to solve a variety of inverse problems:

    We first compare to the gradient descent network \textbf{(GDN)}, an {\em unrolled optimization} algorithm that is trained end-to-end. While theoretical properties GDN and NN are examined in Section~\ref{sec:theory}, we hope to compare the qualitative and quantitative differences between the two architectures. 
   For a fair comparison, we use an identical architecture for the nonlinear learned component in the gradient descent network and the nonlinear learned component in the Neumann network.
   
   We also compare to MOdel-based reconstruction with Deep Learned priors \textbf{(MoDL)} \cite{aggarwal2018modl}, another unrolled algorithm containing a novel data-consistency step that performs conjugate gradient iterations inside the unrolled algorithm. MoDL is also trained end-to-end, and also shares learned parameters between the learned algorithm. In our main experimental section, we use an identical architecture for the learned component of MoDL as is used in GDN and NN.
   
   Trainable Nonlinear Reaction Diffusion \textbf{(TNRD)} \cite{chen2017trainable} is an unrolled optimization algorithm that closely resembles GDN, but with a specific, novel architecture for the learned component motivated by insights from diffusion methods for inverse problems. The learned components in each block consist of a single filter, followed by a learned nonlinearity, and then the transpose of the single filter is applied. Weights are \emph{not} shared across blocks in TNRD. 
   
    The residual autoencoder \textbf{(ResAuto)}, first proposed in \cite{mao2016image}, is an {\em agnostic} method. In Section~\ref{sec:previous}, we discussed agnostic methods that learn a mapping from $\by$ to $\bbeta$, but in these experiments, we consider a variant of an agnostic learner that learns a mapping from $\bX^\top \by$ to $\bhat$ but does not otherwise use $\bX$.  Specifically, we construct a 12-layer convolutional-deconvolutional residual neural network (almost twice as many layers as the network used in the Neumann network), with a channel-wise fully connected layer.
    
    Compressed Sensing using Generative Models, \textbf{(CSGM)} \cite{bora2017compressed} is a {\em decoupled} method which first trains a generative model for the data. After training the generative model, arbitrary inverse problems can be solved by finding the image in the range of the generator which is closest to the distorted image. As in the setup of \cite{bora2017compressed}, in our experiments we train three generative networks, one for each dataset.
    
   Our final method does not incorporate training data at all into the solution of the inverse problem. We reconstruct using total-variation regularized least squares \textbf{(TV)}. We minimize our objective using the algorithm of \cite{wang2008new}, with hyperparameters chosen via cross-validation over a held-out validation set for each dataset and inverse problem.
   
\subsection{Training and Implementation}\label{sec:training_details}

Given training pairs $\{(\bbeta_i,\by_i)\}_{i=1}^N$, and 
assuming the learned component $R$ inside the Neumann network depends smoothly on a set of parameters $\btheta$, \ie the partial derivatives $\partial_{\btheta}R(\bbeta;\btheta)$ exist, we train a Neumann network  $\bhat$ by minimizing the empirical risk $\mathcal{L}(\btheta) = \sum_{i=1}^N \|\bhat(\by_i;\btheta)-\bbeta_i\|^2$.

In the Supplemental Materials we derive the backpropagation gradients $\partial_{\btheta}\mathcal{L}(\btheta)$ in the case where $\bhat$ is a Neumann network or a gradient descent network.

 The learned components of NN, GDN, and MoDL have identical architectures: a 7-layer convolutional-deconvolutional neural network with a single channel-wise fully-connected layer \cite{pathak2016context}, inspired by architectural choices in \cite{dong2016image, mao2016image, kim2016accurate}.

For NN, GDN, MoDL, and TNRD we used architectures with $B = 6$ blocks. The learned component is fixed per network, \ie the learned component in the first block has identical weights to the learned component in all other blocks in any given method and inverse problem, except in TNRD. While using a larger $B$ is possible, we found that increasing $B$ beyond 8 led to greatly increased sensitivity to SGD step size schedule choices. This phenomenon can be observed in Section~\ref{sec:precond}. Anecdotally, we find that it is more difficult to choose SGD step sizes for GDN than for NN even for small $B$, and this difficulty became problematic for $B$ greater than 6.

The ResAuto architecture imitates the architecture of \cite{mao2016image}, an approach that highly resembles the U-Net \cite{ronneberger2015u}, but adjusted for good performance on inverse problems like superresolution, deblurring, and inpainting. Superficially, the architecture resembles an expanded version of the previously-described learned component, with 12 convolution or deconvolution layers instead of 7. Further implementation details can be found in supplementary materials.

\subsection{Small-scale Experiments}

In this section, a variety of methods are used to solve the previously-described inverse problems on three datasets. First, a quantitative comparison in terms of PSNR of the previously-outlined approaches on a variety of datasets and inverse problems is described in Table~~\ref{table:compare_cifar}.

We observe that NN and GDN are competitive across all inverse problems and datasets. State-of-the-art methods like MoDL and TNRD perform quite well across all datasets, but the differences in architecture between PNN and MoDL appear to give an edge to PNN, which we hypothesize is an effect of our previously-highlighted skip connections. All methods that incorporate the forward model into the training and reconstruction process perform competitively in our small-scale experiments.

CSGM appears to suffer because of the lack of training data across all experiments. CSGM must learn the manifold associated with each dataset before being able to produce accurate reconstructions, which in our relatively sample-limited setting appears not to happen. See Figure~\ref{fig:sample_complexity_fig} or the Supplement for examples of images produced by CSGM. While TV reconstructions are reasonably accurate across problems, they are not as accurate as learned approaches, especially in inpainting and compressed sensing.

\begin{table}[htbp]
\centering
\begin{adjustbox}{width=\columnwidth}
\begin{tabular}{ll |lllllll}
& & Inpaint & Deblur & Deblur$+\epsilon$ & CS2 & CS8 & SR4 & SR10 \\ \hline
\parbox[t]{1mm}{\multirow{7}{*}{\rotatebox[origin=c]{90}{CIFAR10}}} & NN & 28.20 & 36.55 & 29.43  & 33.83 & \textbf{25.15} & 24.48 & \textbf{23.09} \\
& PNN & 28.40 & \textbf{37.83} & \textbf{30.47} & 33.75 & 23.43 & \textbf{26.06} & 21.79 \\
& GDN & 27.76 & 31.25 & 29.02 & \textbf{34.99} & 25.00 & 24.49 & 20.47 \\
& MoDL & 28.18 & 34.89 & 29.72 & 33.47 & 23.72 & 24.54 & 21.90 \\
& TNRD & 27.87 & 34.84 & 29.70 & 32.74 & 25.11 & 23.84 & 21.99 \\
& ResAuto & \textbf{29.05} & 31.04 & 25.24 & 18.51 & 9.29 & 24.84 & 21.92 \\
& CSGM & 17.88 & 15.20 & 14.61 & 17.99 & 19.33 & 16.87 & 16.66 \\
& TV & 25.90 & 27.57 & 26.64 & 25.41 & 20.68 & 24.71 & 20.68 \\ \hline \hline
\parbox[t]{1mm}{\multirow{7}{*}{\rotatebox[origin=c]{90}{CelebA}}} & NN & \textbf{31.06} & 31.01 & 30.43  & \textbf{35.12} & \textbf{28.38} & 27.31 & 23.57 \\
& PNN & 30.45 & \textbf{33.79} & \textbf{30.89} & 32.61 & 26.41  & \textbf{28.70} & 23.74 \\
& GDN & 30.99 & 30.19 & 29.27 & 34.93 & 28.33 & 27.14 & 23.46 \\
& MoDL & 30.75 & 30.80 & 29.59 & 30.22 & 25.84 & 26.42 & 24.12 \\
& TNRD & 30.21 & 29.92 & 29.79 & 33.89 & 28.19 & 25.75 & 22.73 \\
& ResAuto & 29.66 & 25.65 & 25.29 & 19.41 & 9.16 & 25.62 & \textbf{24.92} \\
& CSGM & 17.75 & 15.68 & 15.30 & 17.99 & 18.21 & 18.11 & 17.88 \\
& TV & 24.07 & 30.96 & 26.24 & 25.91 & 23.01 & 26.83 & 20.70 \\ \hline \hline
\parbox[t]{1mm}{\multirow{7}{*}{\rotatebox[origin=c]{90}{STL10}}} & NN & 27.47 & 29.43 & 26.12 & \textbf{31.98} & \textbf{26.65} & 24.88 & 21.80 \\
& PNN & 28.00 & \textbf{30.66} & \textbf{27.21} & 31.40 & 23.43 & \textbf{25.95} & \textbf{22.19}  \\
& GDN & \textbf{28.07} & 30.19 & 25.61 & 31.11 & 26.19 & 24.88 & 21.46 \\
& MoDL & 28.03  & 29.42 & 26.06 & 27.29 & 23.16 & 24.67 & 16.88  \\
& TNRD & 27.88 & 29.33 & 26.32 & 31.05 & 25.38 & 24.55 & 21.21 \\
& ResAuto & 27.28 & 25.42 & 25.13 & 19.48 & 9.30 & 24.12 & 21.13 \\
& CSGM & 16.50 & 14.04 & 15.59 & 16.67 & 16.39 & 16.58 & 16.47 \\
& TV & 26.29 & 29.96 & 26.85 & 24.82 & 22.04 & 26.37 & 20.12
\end{tabular}
\end{adjustbox}
\caption{\small PSNR comparison for the CIFAR, CelebA, and STL10 datasets respectively. Values reported are the median across a test set of size 256.}
\label{table:compare_cifar}
\end{table}

\begin{figure}[h]
\centering
\begin{tabular}{c@{}c@{}c@{}c@{}c}
 {\small Original and} \\ \small{$\bX^\top \by$} & {\small NN} & {\small GDN} & {\small ResAuto}  \\ \hline \\[-2ex]
\subfloat{\includegraphics[width = 0.15\textwidth]{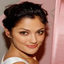}} &
\subfloat{\includegraphics[width = 0.15\textwidth]{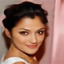}} &
\subfloat{\includegraphics[width = 0.15\textwidth]{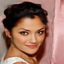}} &
\subfloat{\includegraphics[width = 0.15\textwidth]{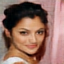}}  \\[-2ex]
\subfloat{\includegraphics[width = 0.15\textwidth]{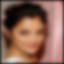}} & \subfloat{\includegraphics[width = 0.15\textwidth]{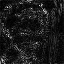}} &
\subfloat{\includegraphics[width = 0.15\textwidth]{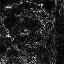}} &
\subfloat{\includegraphics[width = 0.15\textwidth]{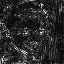}}  \\
\end{tabular}
\caption{\small Reconstruction comparison on the CelebA dataset for the deblur plus noise problem. While the Neumann networks (NN) and gradient descent networks (GDN) perform well, the differences are most apparent the residual images in the second row, especially in the background reconstruction. 
Residuals are formed by displaying the norm across color channels of the error at each pixel, scaled by a factor of 6.}
\label{fig:large_images_cifar_inpaint}
\end{figure}

\begin{figure}[h]
\centering
\begin{tabular}{c@{}c@{}c@{}c@{}c}
 {\small Original and} \\ {\small $\bX^\top \by$} & {\small NN} & {\small GDN} & {\small ResAuto}  \\ \hline \\[-2ex]
\subfloat{\includegraphics[width = 0.15\textwidth]{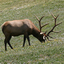}} &
\subfloat{\includegraphics[width = 0.15\textwidth]{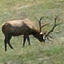}} &
\subfloat{\includegraphics[width = 0.15\textwidth]{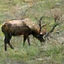}} &
\subfloat{\includegraphics[width = 0.15\textwidth]{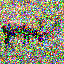}}  \\[-2ex]
\subfloat{\includegraphics[width = 0.15\textwidth]{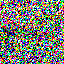}} & \subfloat{\includegraphics[width = 0.15\textwidth]{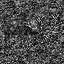}} &
\subfloat{\includegraphics[width = 0.15\textwidth]{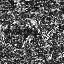}} &
\subfloat{\includegraphics[width = 0.15\textwidth]{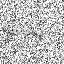}}
\end{tabular}
\caption{\small 8x compressed sensing reconstruction comparison on STL10. ResAuto fails to invert the compressed sensing problem adequately. The Gradient Descent network (GDN) reconstructs accurately but generates more artifacts than the Neumann network (NN).}
\label{fig:large_images_celeba_superres}
\end{figure}

The residual autoencoder in particular has excellent performance on certain problems like inpainting and superresolution, but is not competitive for compressed sensing and deblurring. Recall the motivation for the residual autoencoder: the closer $\bX^\T \by$ is to the ground truth $\bstar$, the simpler the residual $\bstar-\bX^\T \by$ that the network must learn. With this in mind, it seems reasonable that the residual autoencoder should perform well on small-scale downsampling, inpainting, and deblurring, but would fail to generate high-quality reconstructions for compressed sensing or heavy downsampling where $\bX^\T\by$ is likely to be a poor approximation of $\bstar$.

The difference in performance between PNN and NN in Table \ref{table:compare_cifar} can provide some insight regarding the usage of various architectures. First, although PNN performs well for 4x superresolution, deblurring, and deblurring with noise, preconditioning is not a universal solution: inpainting and compressed sensing are perfectly conditioned and preconditioning appears to worsen performance. We see similar effects with MoDL, which performs better than NN or GDN on certain problems, but suffers especially in compressed sensing. These results further emphasize that consideration of the specific forward model at hand should be an important element of designing learned inverse problem solvers.

In addition, we observe some variance in results across datasets. CIFAR10 in particular seems to be a an outlier: while the best reconstructions on CelebA are uniformly more accurate than on STL10, the apparent "difficulty" of reconstruction in CIFAR10 is more task-specific.

Figures~\ref{fig:large_images_cifar_inpaint} and \ref{fig:large_images_celeba_superres} demonstrate more qualitative and quantitative detail in some examples from several different inverse problems and all three datasets. In these figures the residuals are shown for illustrative purposes: the residuals are formed by displaying the scaled pixelwise norm across color channels of the difference $\bstar - \bhat$ where $\bstar$ is the true image, and $\bhat$ the estimate, scaled by a factor of 6. Magnitudes are clipped to be less than or equal to 1.

\subsection{Effect of Sample Size}

In section \ref{sec:decoupled} we hypothesized that incorporating information about the forward operator would have implications for the sample sizes required to achieve particular error rates.

\begin{figure}[ht!]
\centering
\renewcommand{\arraystretch}{0.1}
\begin{tabular}{c|@{}c@{}c@{}c@{}c@{}c}
 & \begin{tabular}{@{}c@{}}{\scriptsize Original and} \\ {\scriptsize $\bX^\top \by$}\end{tabular} & {\scriptsize NN} & {\scriptsize GDN} & {\scriptsize ResAuto} & {\scriptsize CSGM}  \\ \hline \\[-2ex]
 \raisebox{1.5em}{2k} & \subfloat{\includegraphics[width = 0.12\textwidth]{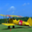}}  & \subfloat{\includegraphics[width = 0.12\textwidth]{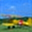}} &
\subfloat{\includegraphics[width = 0.12\textwidth]{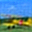}} &
\subfloat{\includegraphics[width = 0.12\textwidth]{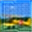}} &
\subfloat{\includegraphics[width = 0.12\textwidth]{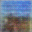}}\\[-2ex]
\raisebox{2em}{\begin{tabular}{@{}c@{}}{\scriptsize $\bX^\top \by$ and} \\ {\scriptsize Residuals}\end{tabular}} & \subfloat{\includegraphics[width = 0.12\textwidth]{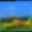}} & \subfloat{\includegraphics[width = 0.12\textwidth]{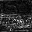}} &
\subfloat{\includegraphics[width = 0.12\textwidth]{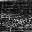}} &
\subfloat{\includegraphics[width = 0.12\textwidth]{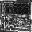}} & \subfloat{\includegraphics[width = 0.12\textwidth]{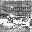}} \\[-2ex]
\raisebox{1.5em}{30k} & \subfloat{\includegraphics[width = 0.12\textwidth]{figures/sample_comp/30k/orig}} & \subfloat{\includegraphics[width = 0.12\textwidth]{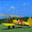}} &
\subfloat{\includegraphics[width = 0.12\textwidth]{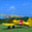}} &
\subfloat{\includegraphics[width = 0.12\textwidth]{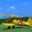}} &
\subfloat{\includegraphics[width = 0.12\textwidth]{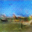}}\\[-2ex]
\raisebox{2em}{\begin{tabular}{@{}c@{}}{\scriptsize $\bX^\top \by$ and} \\ {\scriptsize Residuals}\end{tabular}} & \subfloat{\includegraphics[width = 0.12\textwidth]{figures/sample_comp/30k/altered}} & \subfloat{\includegraphics[width = 0.12\textwidth]{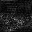}} &
\subfloat{\includegraphics[width = 0.12\textwidth]{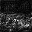}} &
\subfloat{\includegraphics[width = 0.12\textwidth]{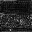}} & \subfloat{\includegraphics[width = 0.12\textwidth]{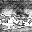}} \\
\end{tabular}
\caption{\small A qualitative comparison of the reconstructions produced for the deblurring problem on a single image at two different training set sizes, along with the associated residual images. Residual images are scaled by a factor of 6.}
\label{fig:sample_complexity_fig}
\end{figure}%

\begin{figure}[ht!]
\centering
\subfloat[Sample Complexity]{\includegraphics[width = 0.4\textwidth]{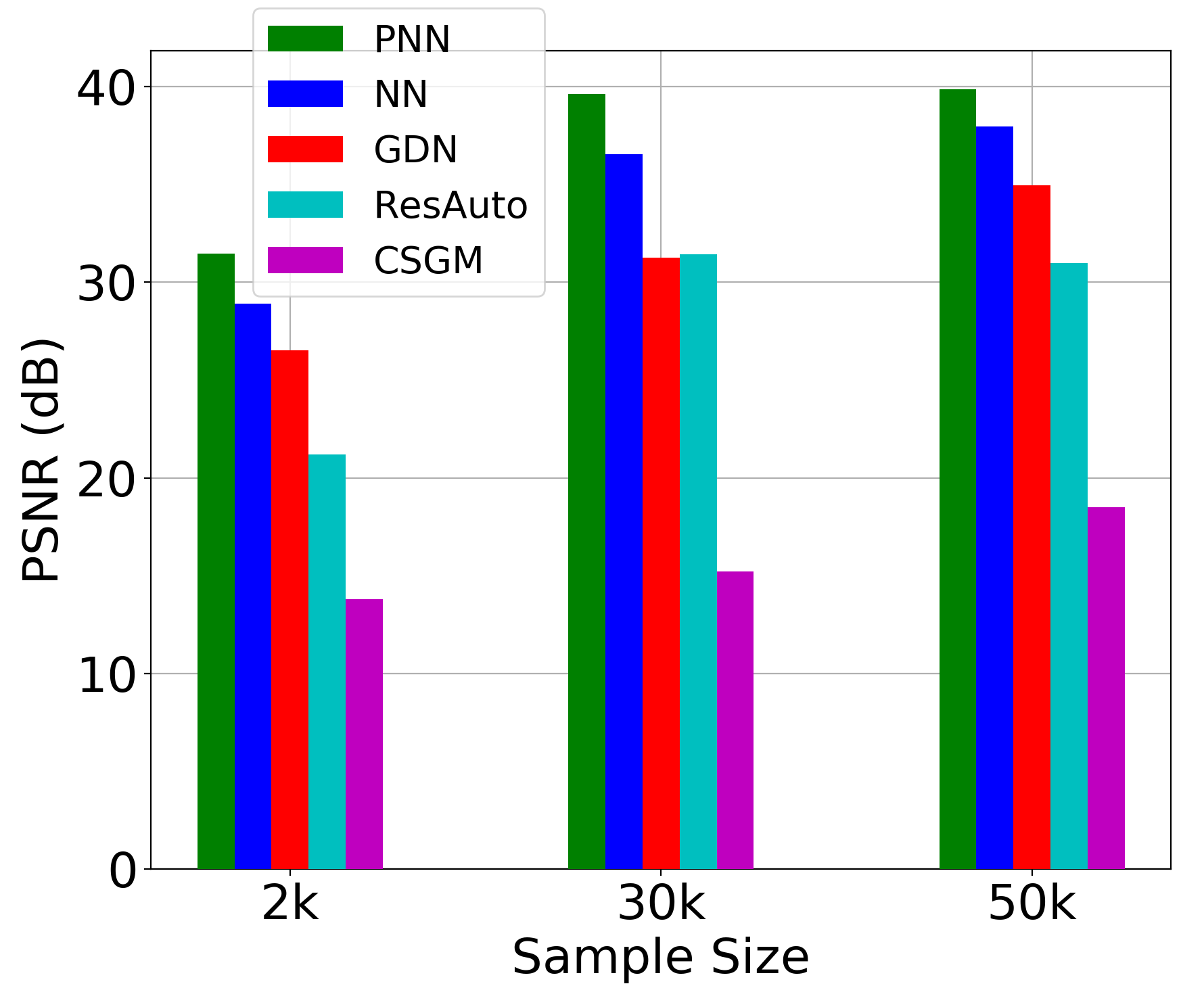}\label{fig:sample_complexity_bar}}~~
\subfloat[Preconditioning]{\includegraphics[width = 0.4\textwidth]{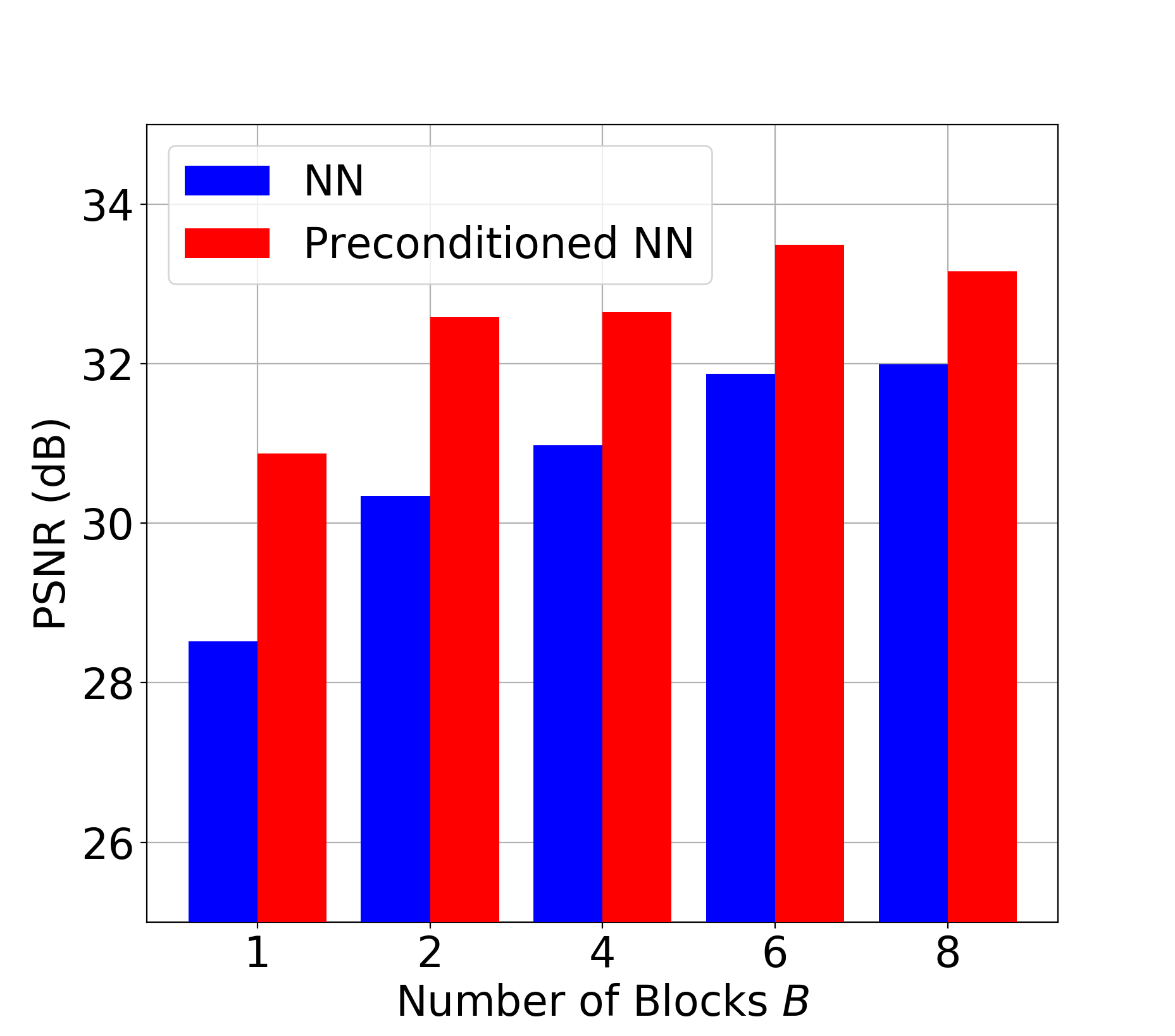} \label{fig:precondition_bar}}
\caption{\small Performance comparisons. (a) Median PSNR of methods trained with different sample sizes of 2,000, 30,000, and 50,000. Neumann networks (NN) and Preconditioned Neumann networks (PNN) scale very well with training set size, with smaller marginal gains as training sizes increase. All PSNR values are for the CIFAR-10 dataset, and the inverse problem used is the previously described deblurring problem. (b) PSNR (dB) for the standard and preconditioned NN. The inverse problem in this case is deblurring with a Gaussian kernel of size $5 \times 5$ and variance $\sigma = 5.0$.}
\end{figure}%

A coarse comparison of the presented learning-based  methods at different sample sizes is provided in Figure~\ref{fig:sample_complexity_bar}. We observe that while all methods suffer a decrease in PSNR at low sample sizes, the Neumann network  has the highest-quality reconstructions at only 2,000 images, and also enjoys the largest increase of performance when going from 2,000 to 30,000 training images. Gradient descent network performs well even at very low sample sizes, but artifacts are present in reconstructions at low sample sizes, visible in Figure~\ref{fig:sample_complexity_fig}.

Methods that do not incorporate the forward model, like ResAuto and CSGM, perform poorly in the low-sample regime, as discussed in Section~\ref{sec:decoupled}. While ResAuto performs competitively at 30k iterations, there is little change between image qualities produced at these sample sizes, and even a very slight decrease in performance. CSGM improves significantly with increasing samples, but does not produce high-quality reconstructions on this inverse problem.

\subsection{Effect of Preconditioning}\label{sec:precond}

Figure~\ref{fig:precondition_bar} illustrates the effect of preconditioning on the performance of the Neumann network with different numbers of blocks $B$ on a deblurring task. While the original Neumann network does not surpass 32 dB PSNR with 8 blocks, the preconditioned Neumann network surpasses the original with only $B = 2$, and continues to improve as the number of blocks increases. Example images are included in the supplementary materials.

The forward problem in this case is Gaussian deblurring with $\sigma = 5.0$ and a blur kernel of size $5 \times 5$. The corresponding $\bX$ is very poorly conditioned, and a $\lambda$ of 0.01 is used in the preconditioning matrix $(\bX^\top \bX + \lambda \bI)^{-1}$.

Depending on the structure of $\bX$ and how easily $(\bX^\top \bX + \lambda \bI)^{-1}$ can be computed, preconditioning can be computationally costly, but it appears to permit fewer Neumann network blocks for comparable performance. Since the primary resource bottleneck for training the Neumann network end-to-end is memory, fewer blocks permits faster training, or alternately, allows implementations to achieve higher performance than would otherwise be possible with fixed computational resources.

\subsection{MRI Experiments}\label{sec:mri}

\begin{figure*}[ht!]
\centering
\renewcommand{\arraystretch}{0.1}
\begin{tabular}{@{}c@{}c@{}c@{}c@{}c@{}c@{}c@{}c}
\subfloat{\includegraphics[width = 0.12\textwidth]{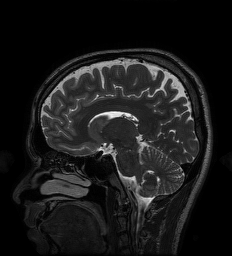}}  & 
\subfloat{\includegraphics[width = 0.12\textwidth]{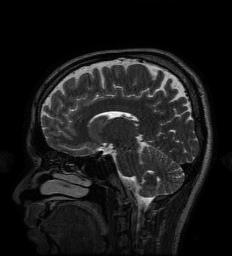}} &
\subfloat{\includegraphics[width = 0.12\textwidth]{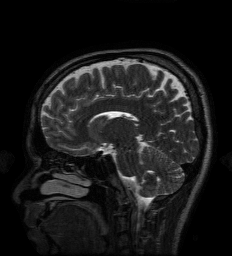}} &
\subfloat{\includegraphics[width = 0.12\textwidth]{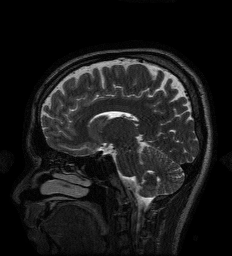}} &
\subfloat{\includegraphics[width = 0.12\textwidth]{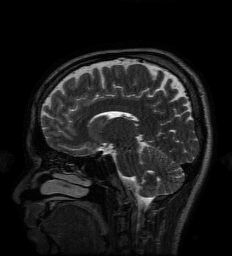}} &
\subfloat{\includegraphics[width = 0.12\textwidth]{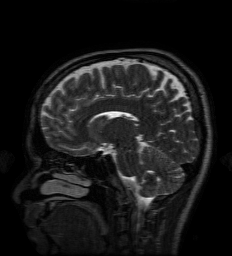}} & 
\subfloat{\includegraphics[width = 0.12\textwidth]{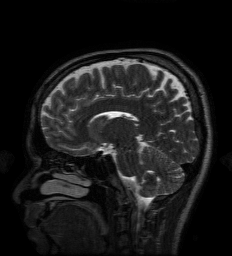}} & 
\subfloat{\includegraphics[width = 0.12\textwidth]{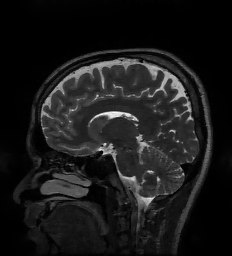}} \\[-2ex] 
\subfloat{\includegraphics[width = 0.12\textwidth]{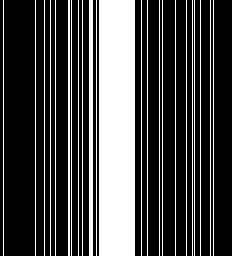}} &
\subfloat{\includegraphics[width = 0.12\textwidth]{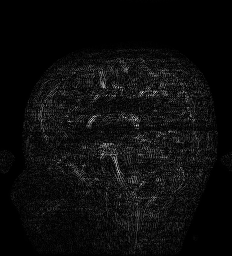}} &
\subfloat{\includegraphics[width = 0.12\textwidth]{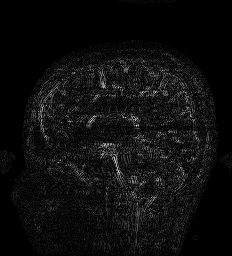}} &
\subfloat{\includegraphics[width = 0.12\textwidth]{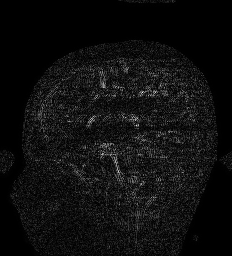}} &
\subfloat{\includegraphics[width = 0.12\textwidth]{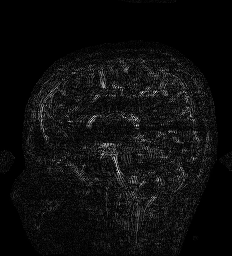}} & 
\subfloat{\includegraphics[width = 0.12\textwidth]{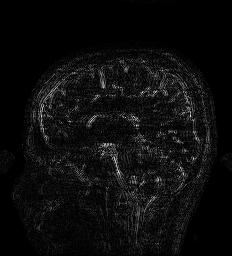}} &
\subfloat{\includegraphics[width = 0.12\textwidth]{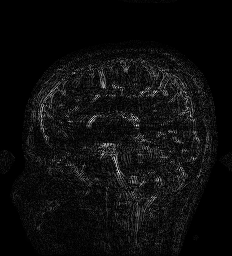}} & 
\subfloat{\includegraphics[width = 0.12\textwidth]{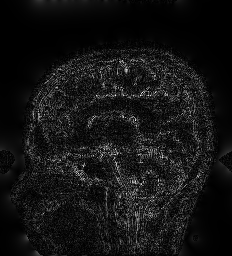}} \\ \\ 
{\scriptsize Original/Mask } & {\scriptsize PNN} 
& {\scriptsize NN}
& {\scriptsize MoDL} & {\scriptsize GDN2} & {\scriptsize GDN1}  & {\scriptsize TNRD} 
& {\scriptsize TV}  \\ \\
{\scriptsize PSNR (dB) } & {\scriptsize 34.95 dB} 
& {\scriptsize 33.09 dB}
& {\scriptsize 34.09 dB} & {\scriptsize 33.18 dB} & {\scriptsize 31.37 dB}  & {\scriptsize 32.39 dB} 
& {\scriptsize 32.29 dB} \\ \\
{\scriptsize Test Time (sec) } & {\scriptsize 16.3 sec} 
& {\scriptsize 5.5 sec}
& {\scriptsize 14.3 sec} & {\scriptsize 5.7 sec} & {\scriptsize 3.1 sec}  & {\scriptsize 4.0 sec} 
& {\scriptsize 349.2 sec} 
\end{tabular}
\caption{\small A comparison of MRI reconstruction quality for a variety of trainable and non-trainable image reconstruction methods. The 12-coil data is undersampled by a factor of $4\times$ and Gaussian noise with $\sigma=0.01$ is added in k-space. The reconstructions are displayed in the first row, while the second row contains the residual images scaled by a factor of $4$. PSNR is displayed next to the method name, while below each method name is the mean time required to reconstruct a single MRI image in seconds. GDN2 denotes the Gradient Descent network using the same initialization as the preconditioned Neumann network, while GDN1 uses the same initialization as the Neumann network.}
\label{fig:mri_reconstruction}
\end{figure*}%

In this section we provide results of multi-coil MRI reconstruction from undersampled measurements. Full training and test data is the data used for the experiments in \cite{aggarwal2018modl}, consisting of 12-coil Cartesian sampled k-space data of dimension $232 \times 208 \times 12$ with known coil sensitivity maps. The size of the training set is 360 such acquisitions across 4 subjects, with testing being performed on 40 images from one, separate subject who was not used for training. The sum-of-squares reconstruction is treated as ground truth. Further details of the data acquisition can be found in \cite{aggarwal2018modl}.

All experiments are for 4$\times$ undersampling, although we differ from \cite{aggarwal2018modl} in that we train on a fixed k-space undersampling mask. The undersampling mask is fully sampled in the center 0.15 fraction of frequencies, with the remaining frequencies being sampled according to a random Gaussian pattern. The mask is visualized in figure \ref{fig:mri_reconstruction}.

For the MRI experiments we follow the precedent set by \cite{aggarwal2018modl} in our choice of learned component, using only a simple five-layer convolutional network with 64 filters per layer and ReLU nonlinearities for all architectures other than TNRD. The TNRD architecture follows the architecture proposed in \cite{chen2017trainable}. The Neumann network results presented here are for the preconditioned Neumann network (PNN), and the number of blocks for GDN, PNN, MoDL, and TNRD is fixed to be 5. The preconditioning operator in PNN is implemented through 10 conjugate gradient iterations, identically to \cite{aggarwal2018modl}. We compare to GDN with the same initialization as NN (GDN1) and as PNN (GDN2) to study the effect of different initializations on GDN.

We observe that unrolled optimization approaches are advantageous in this setting compared to the more traditional TV-regularized reconstruction. Preconditioning, both to improve initialization as in GDN2, and incorporated into the architectures, as in PNN and MoDL, improves PSNR significantly in this setting.

A major benefit of learned reconstruction methods is their test time, which is displayed beneath the method name and PSNR in Figure \ref{fig:mri_reconstruction}. We note that all learned approaches reconstruct an order of magnitude faster than the agnostic TV approach. Although preconditioning incurs an additional cost in terms of test time, the performance increase is substantial for MoDL and PNN.

\subsection{Optimization Landscapes}\label{sec:optlandscape}

\begin{figure}[htpb]
\centering
\begin{tabular}{c@{}c@{}c@{}c@{}c}
\subfloat{\includegraphics[width = 0.2\textwidth]{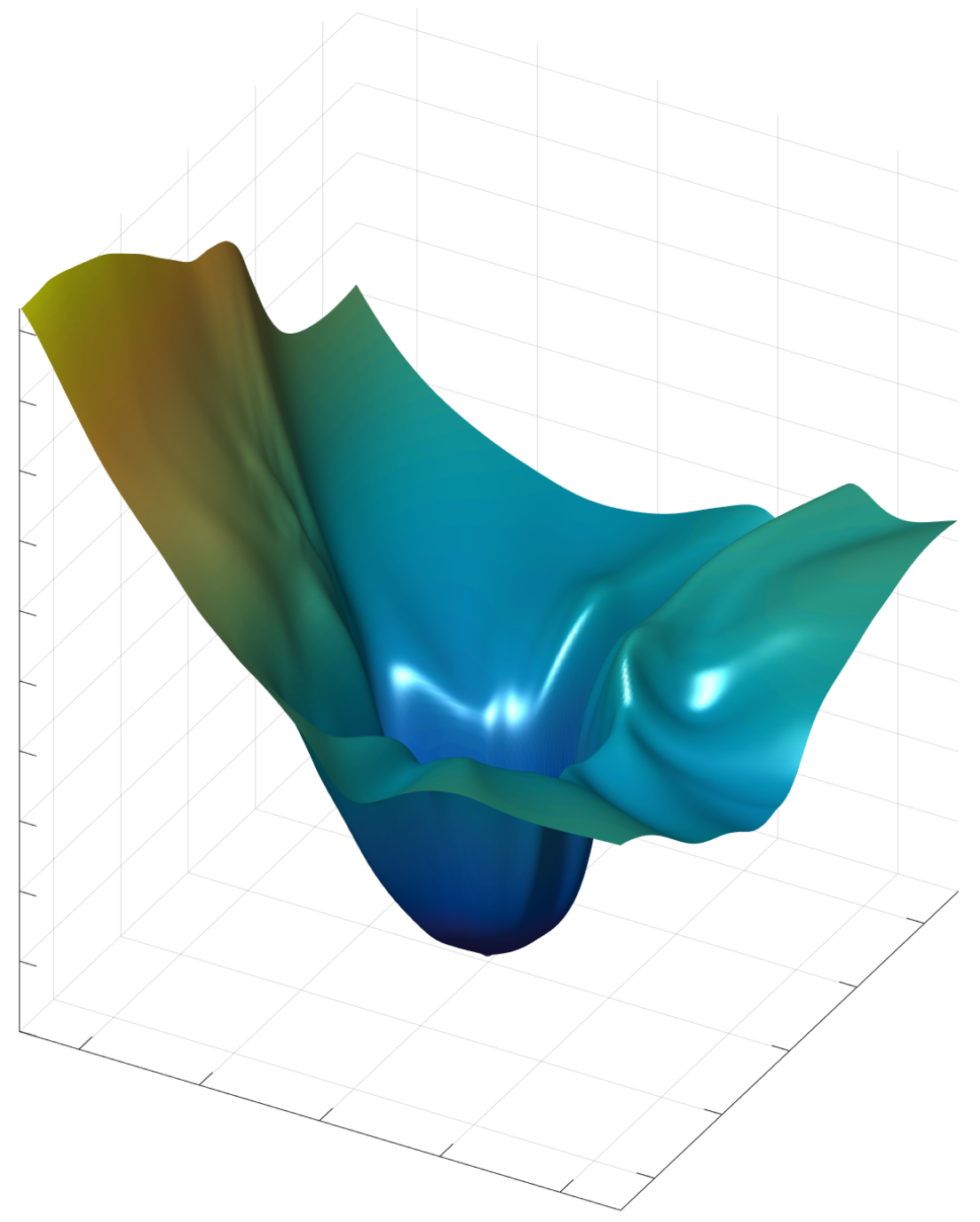}} &
\subfloat{\includegraphics[width = 0.2\textwidth]{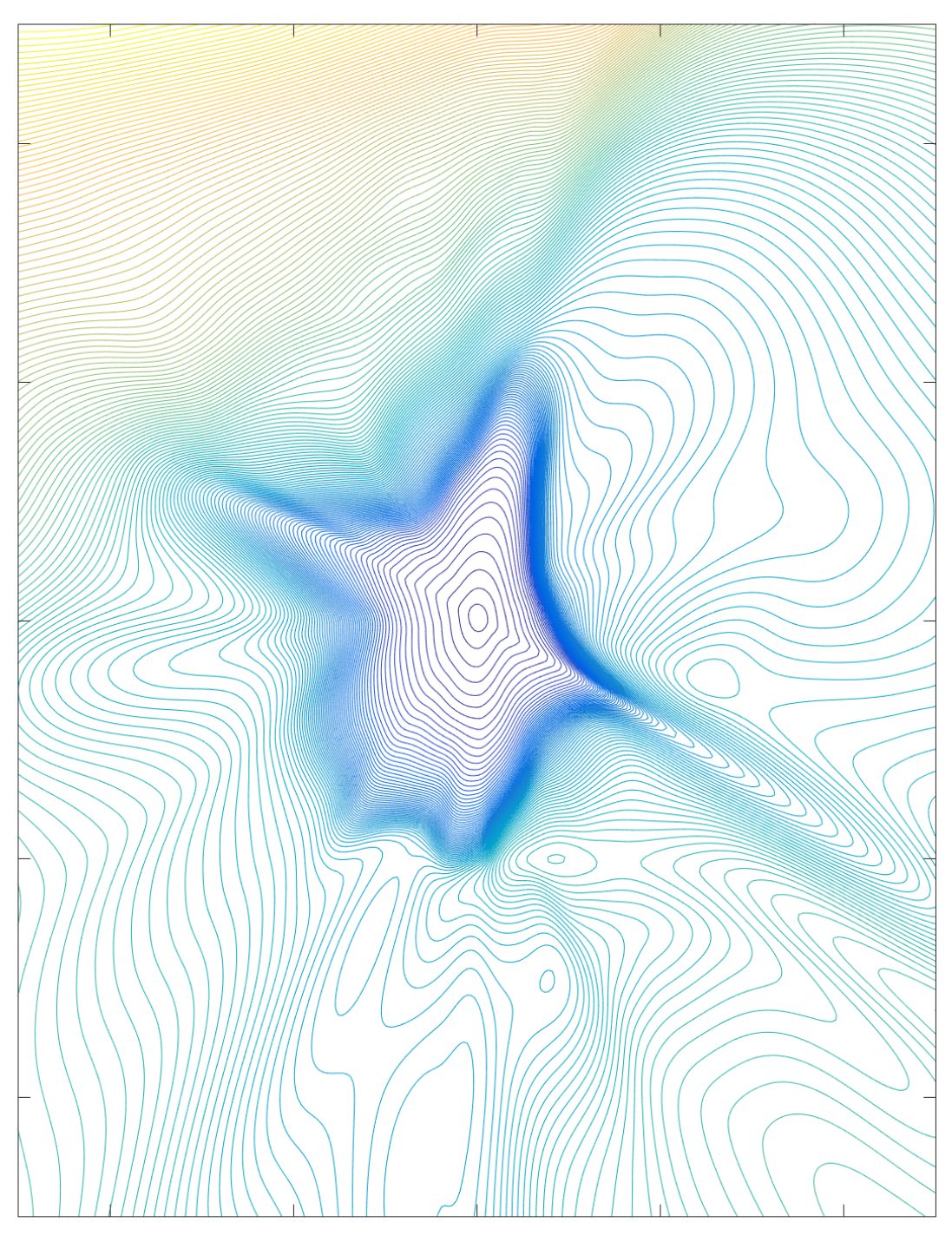}} &
\subfloat{\includegraphics[width = 0.2\textwidth]{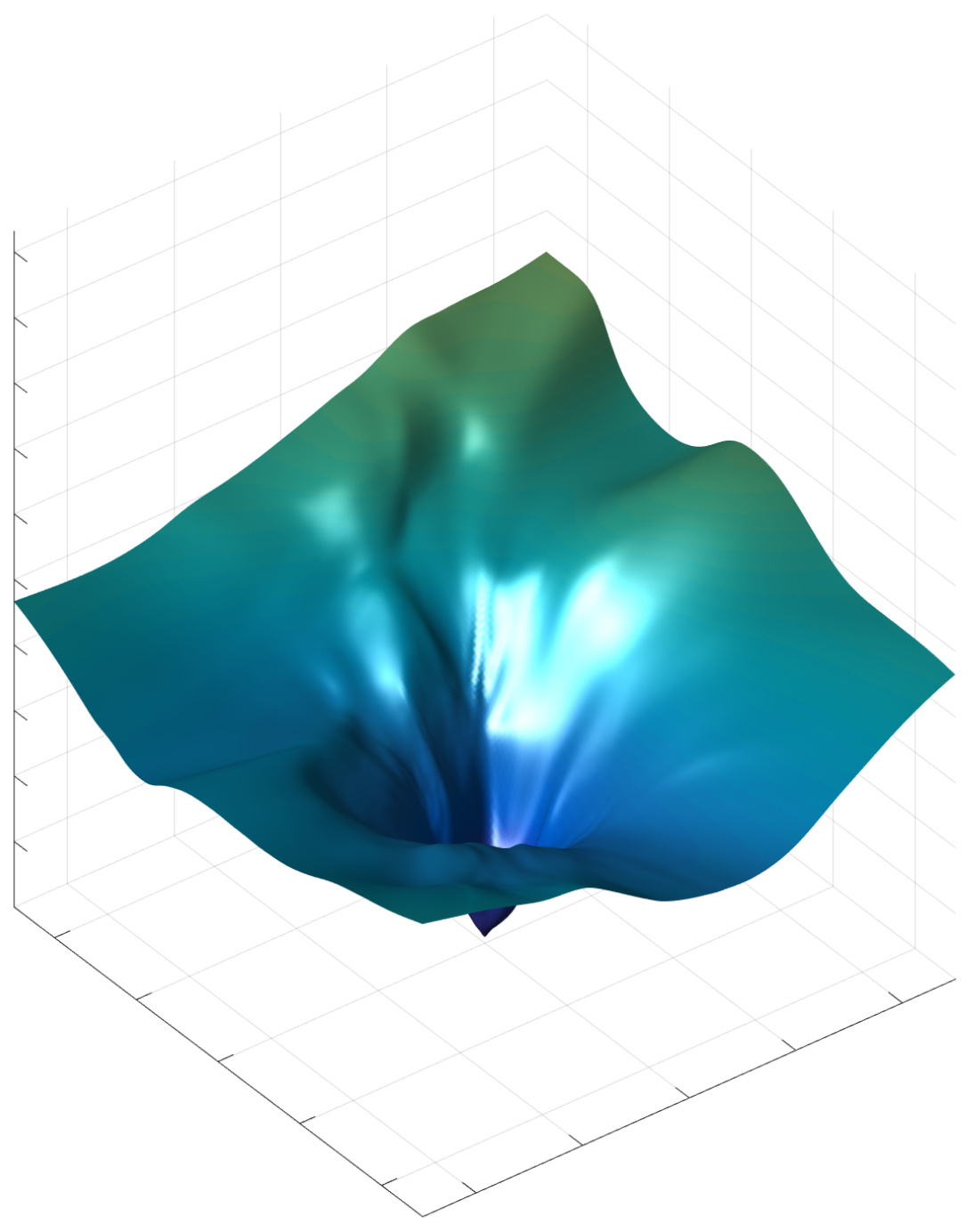}} &
\subfloat{\includegraphics[width = 0.2\textwidth]{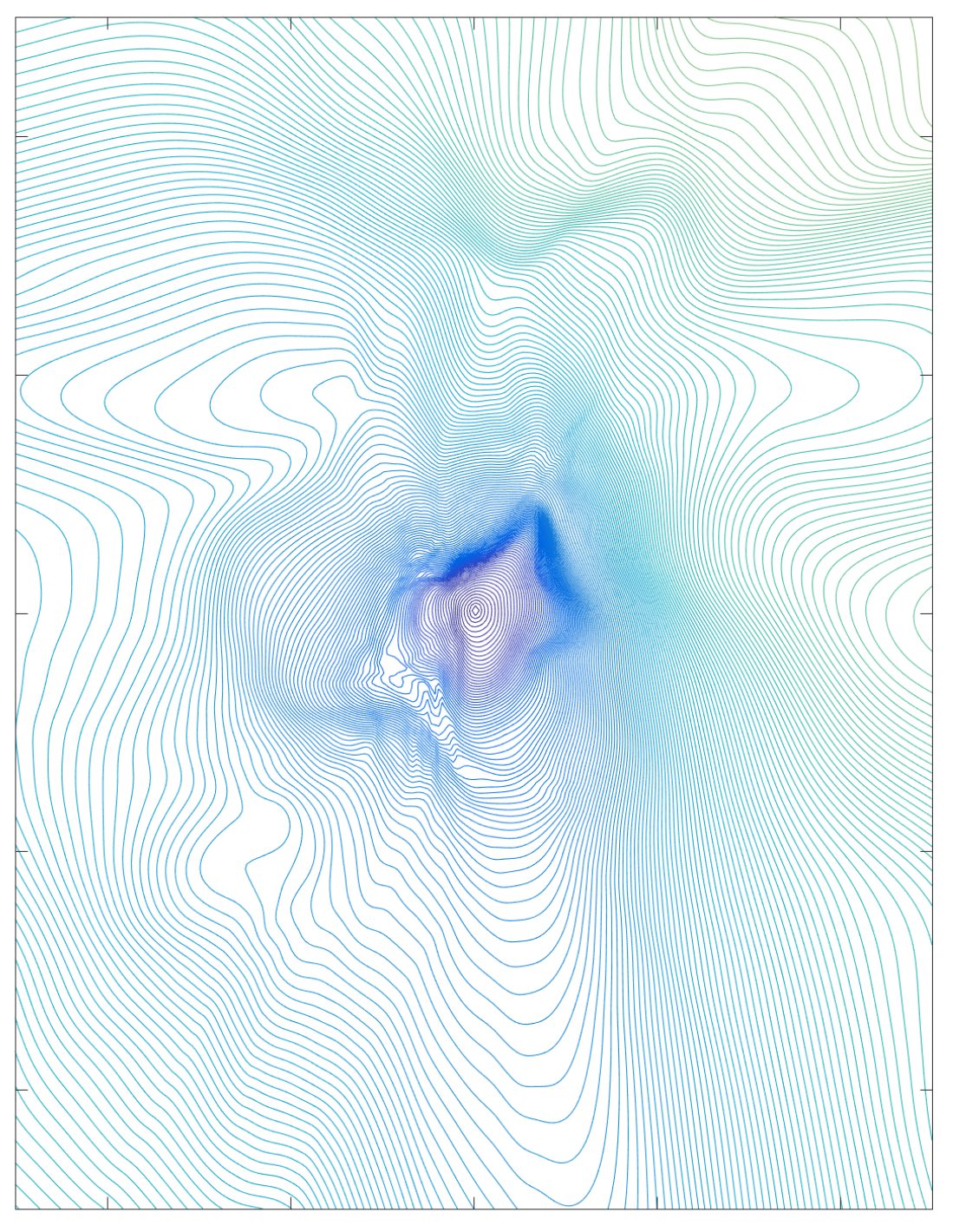}}  \\[-2ex]
\subfloat{\includegraphics[width = 0.2\textwidth]{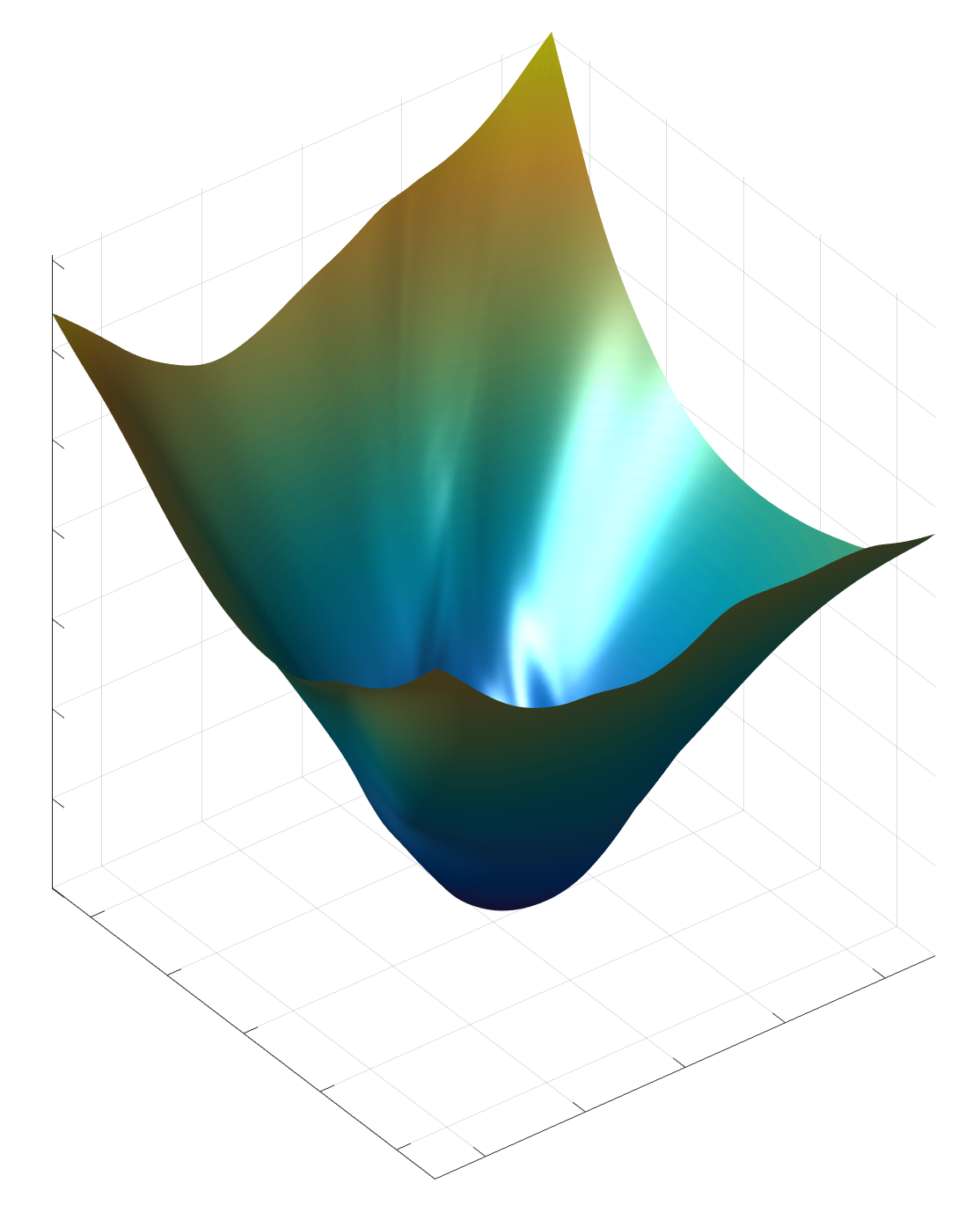}} &
\subfloat{\includegraphics[width = 0.2\textwidth]{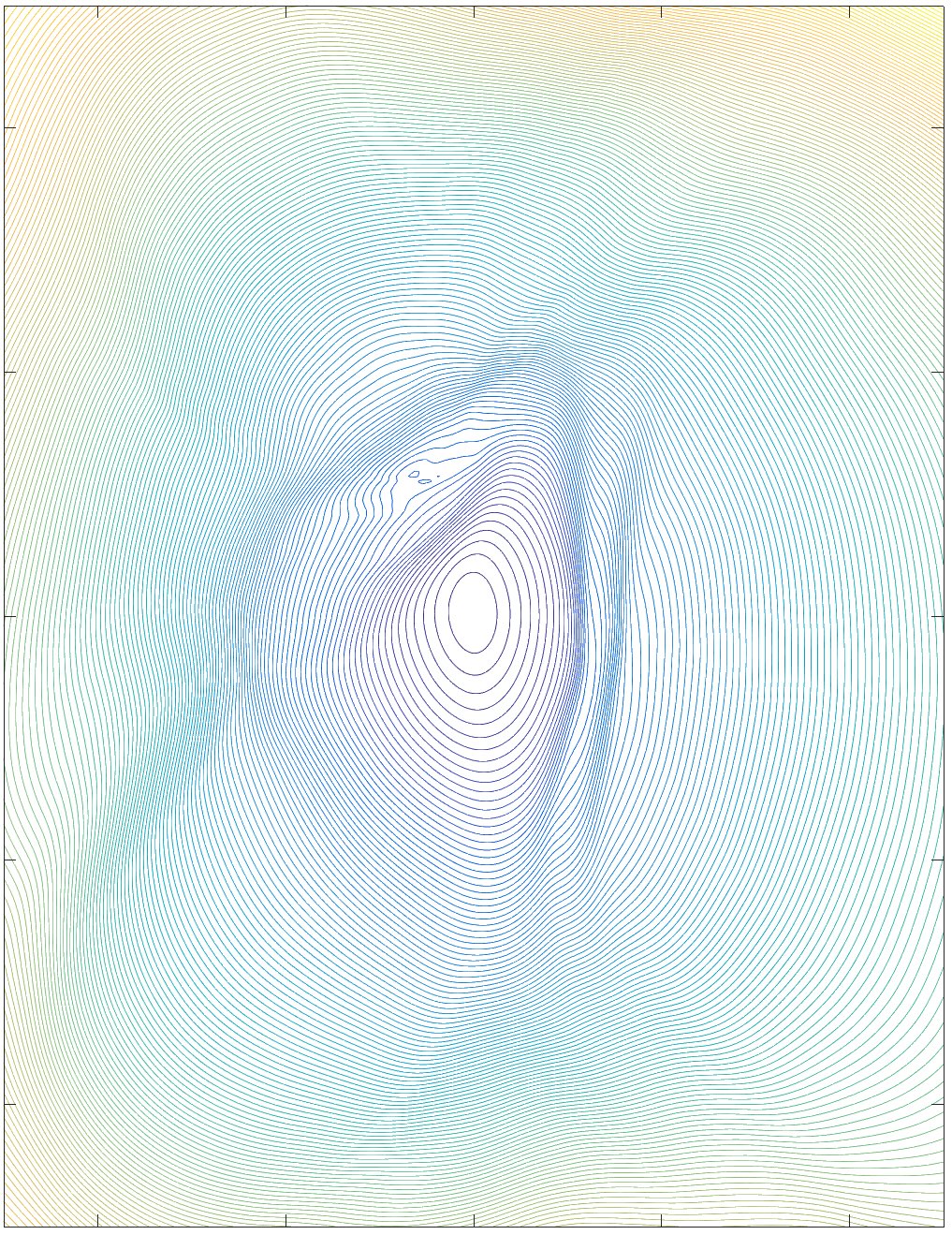}} &
\subfloat{\includegraphics[width = 0.2\textwidth]{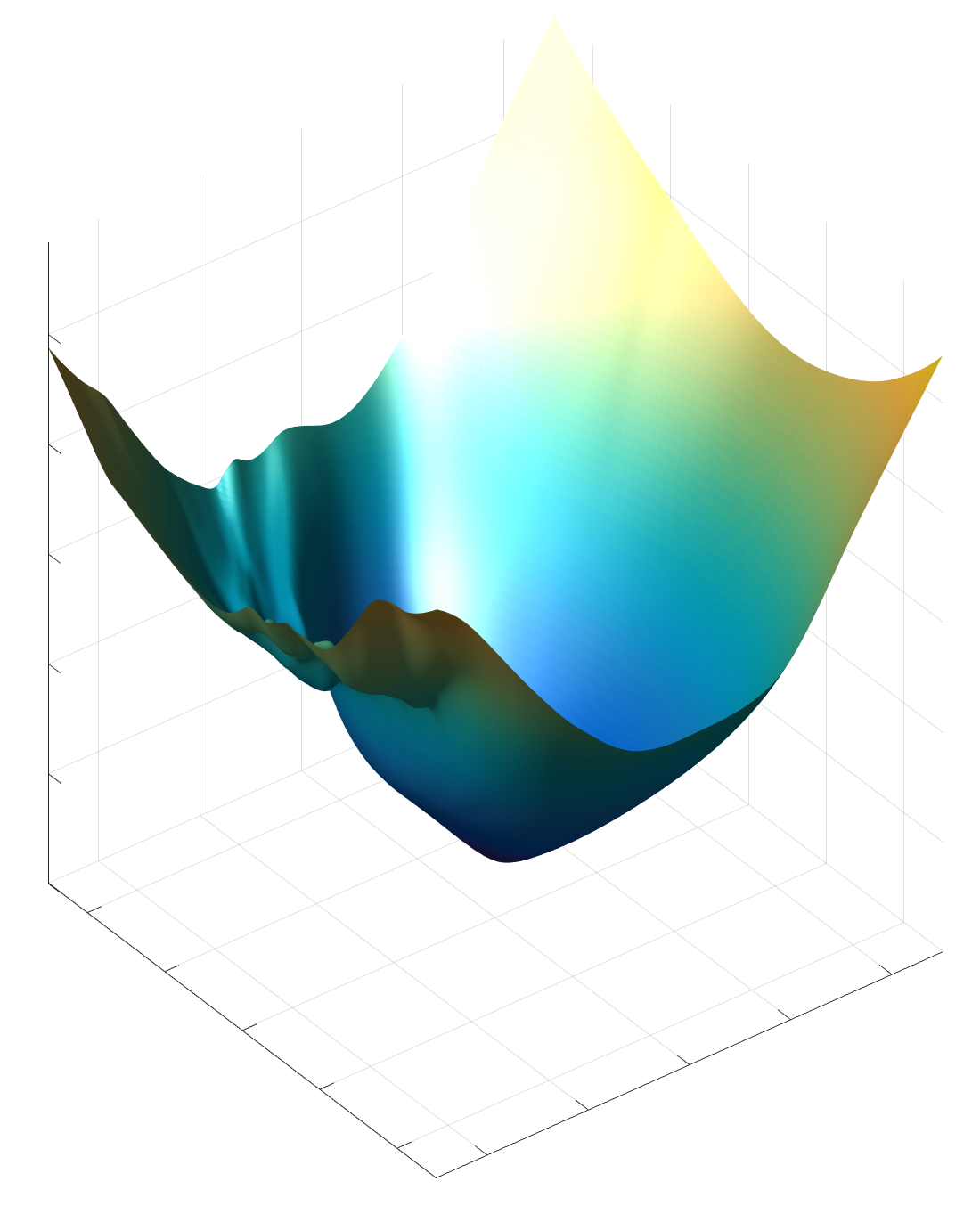}} &
\subfloat{\includegraphics[width = 0.2\textwidth]{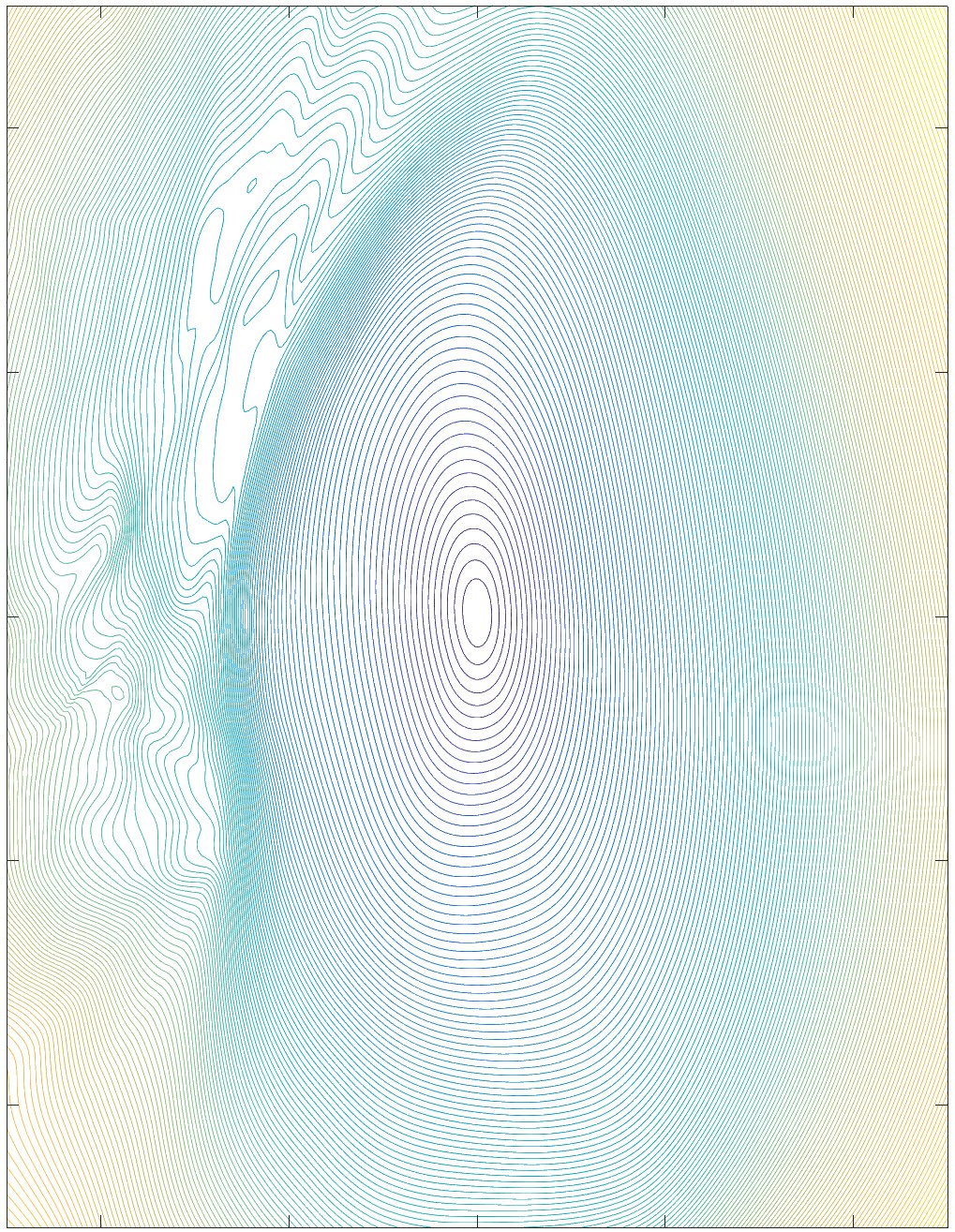}}  \\[-2ex]
\subfloat{\includegraphics[width = 0.2\textwidth]{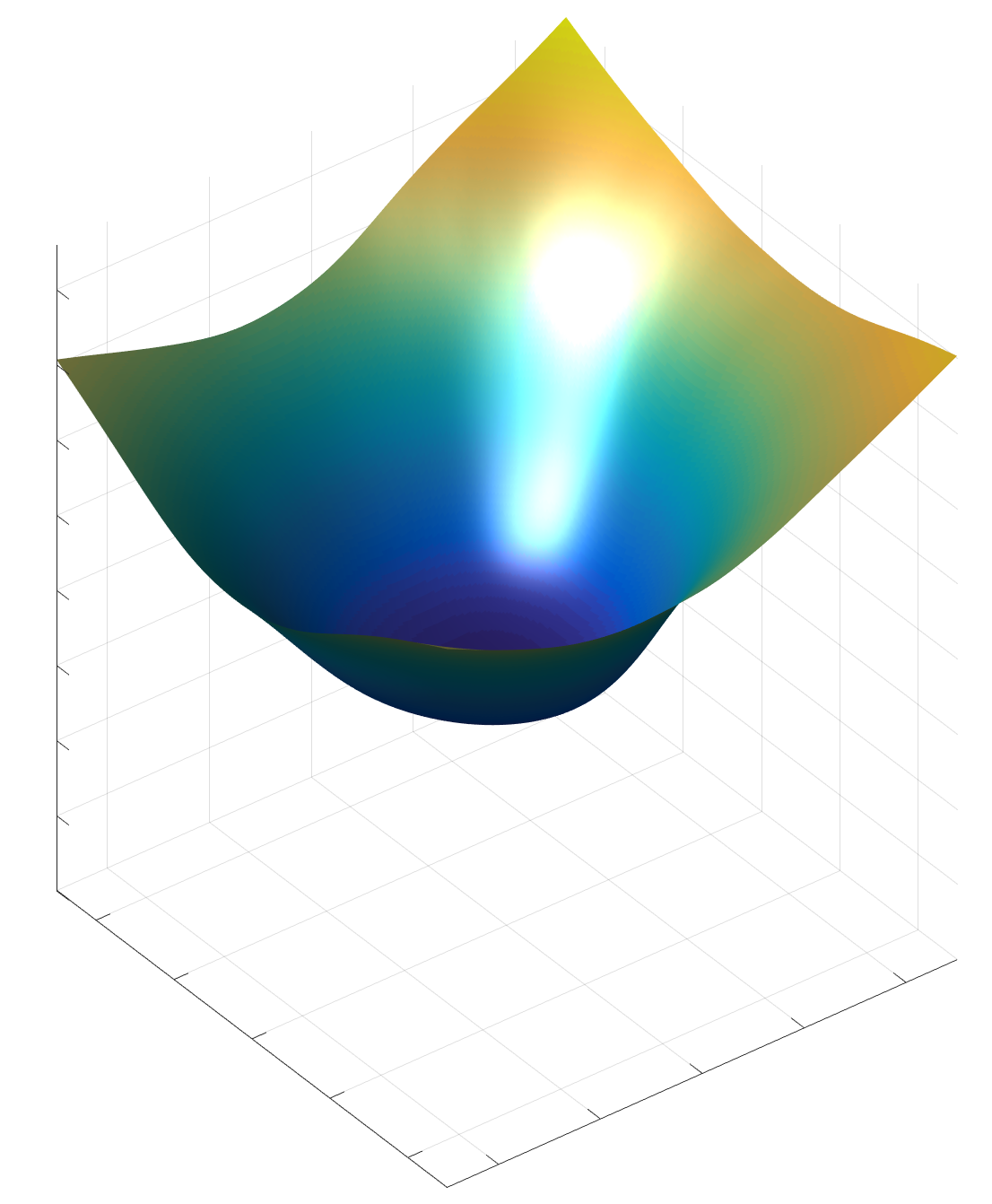}} & \subfloat{\includegraphics[width = 0.2\textwidth]{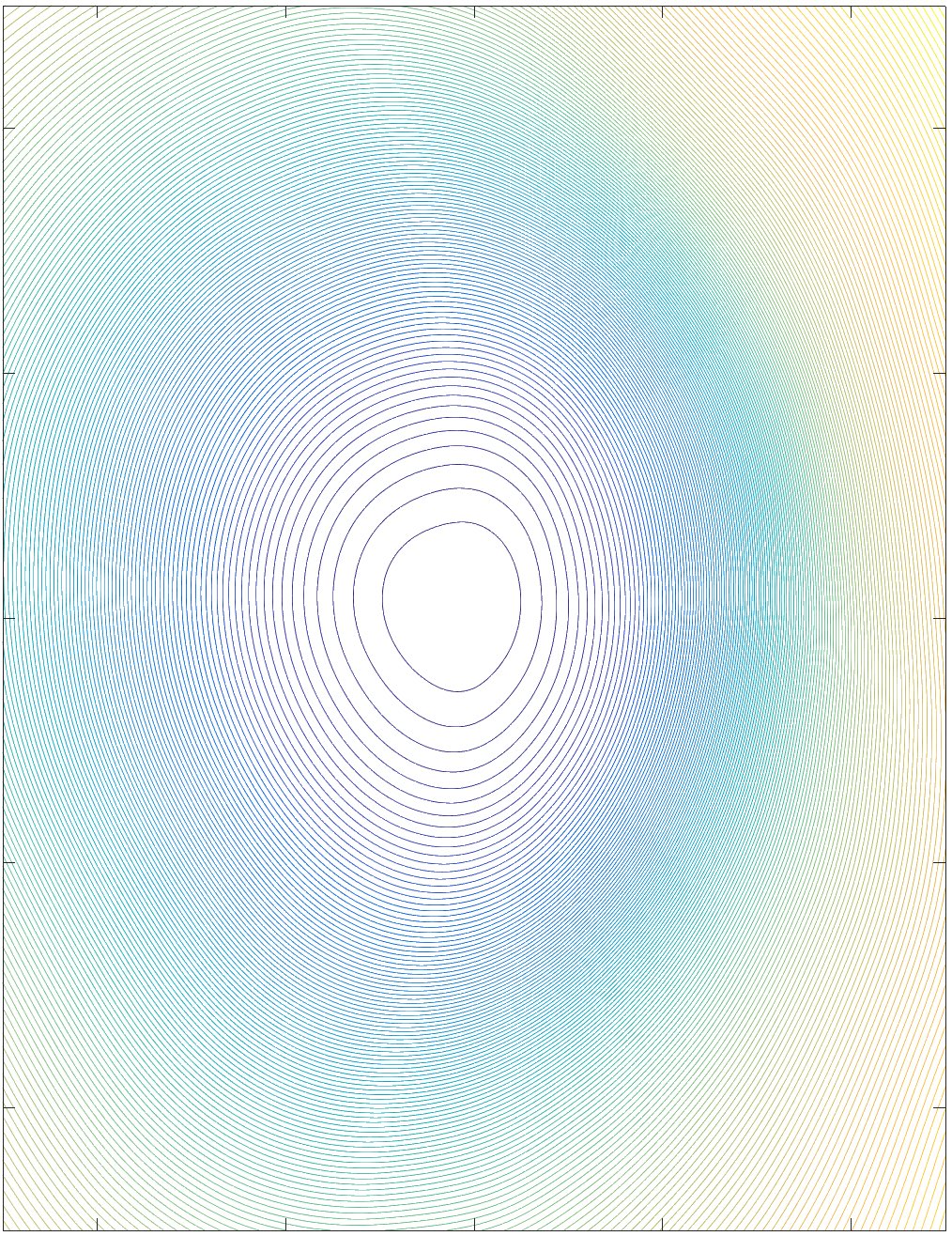}} &
\subfloat{\includegraphics[width = 0.2\textwidth]{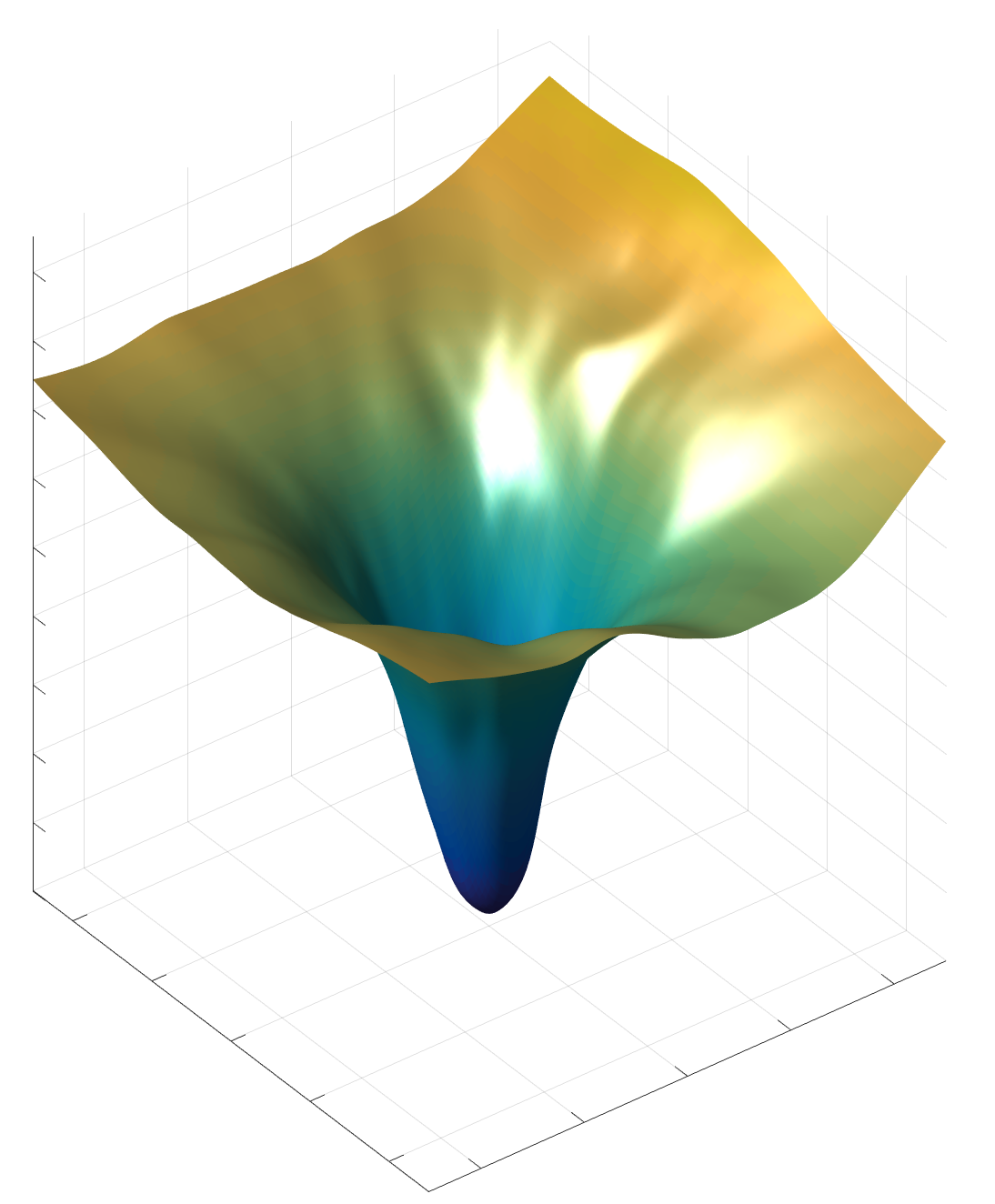}} &
\subfloat{\includegraphics[width = 0.2\textwidth]{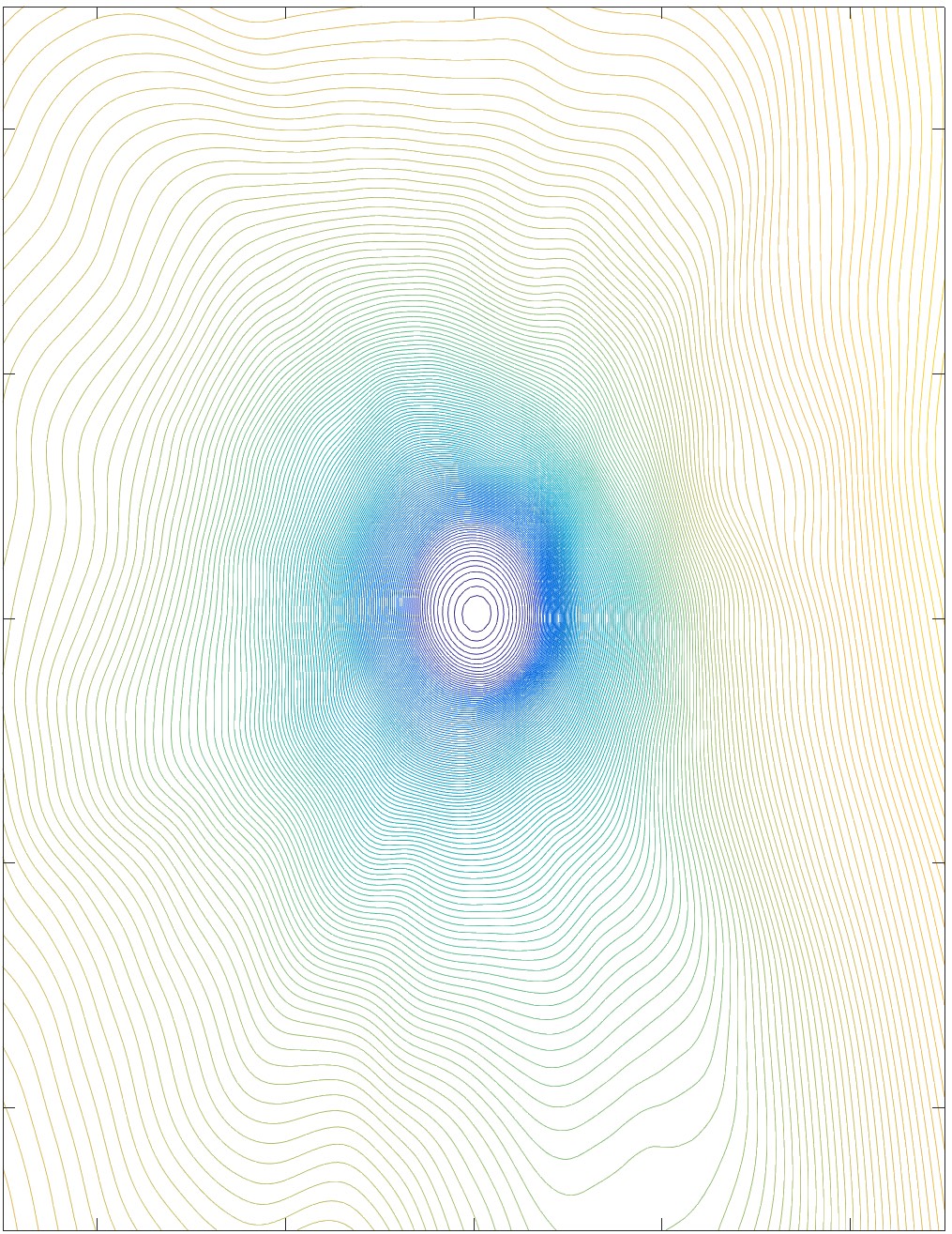}}  \\
  {\begin{tabular}{@{}c@{}}{\scriptsize (a) Neumann Loss} \\[-0.5em] {\scriptsize Surface}\end{tabular}} &  {\begin{tabular}{@{}c@{}}{ \scriptsize (b) Neumann} \\[-0.5em] {\scriptsize Contour}\end{tabular}} & {\begin{tabular}{@{}c@{}}{\scriptsize (c) GDN Loss} \\[-0.5em] {\scriptsize Surface}\end{tabular}} & {\scriptsize (d) GDN Contour}  
\end{tabular}
\caption{\small Optimization landscapes and contour plots. (a) The optimization landscape associated with the training loss of the Neumann network around the center optimal point. (b) The associated contour plot. (c) The optimization landscape associated with the training loss of the gradient descent network. (d) The associated contour plot. 
Neumann network landscapes tend to have wider basins around the minimizer and be steeper outside the basin of the minimizer, which are both more favorable to practical optimization by SGD.
Figures use the CIFAR-10 dataset and, from top to bottom, deblurring inverse problem with $\sigma=2.5$, $10\times$ superresolution, and compressed sensing with $8\times$ compression.}
\label{fig:opt_landscape_test}
\end{figure}

The performance of the Neumann networks (NN) and Gradient Descent networks (GDN) are very similar across a range of problems and datasets, but NN slightly outperforms GDN consistently. We hypothesize this is due to differences in the connectivity of their network architectures and the effect this has on training.

Specifically, both NN and GDN contain connections across blocks, but differ mainly in their \emph{direction} and \emph{extent}. Adjacent blocks in both networks share residual connections as in a ResNet \cite{he2016deep} (the inclusion of the identity $\bI$ in the linear part $[\bI-\eta\bX^\T\bX](\cdot)$ of each block of NN and GDN is a residual connection). However, the main difference is that NN contain additional ``skip'' connections that connect each block with the final layer, similar to architectures like DenseNets \cite{huang2017densely}. Recent work \cite{li2018visualizing} has highlighted the role of residual connections in the optimization landscape of deep architectures, implying that residual connections ``smooth" the optimization landscape. Specifically, fewer local minima tend to be present, and those minima tend to be wide, as opposed to sharp. In addition, the authors of \cite{li2018visualizing} note that skip connections from intermediate or early layers of deep networks to final layer tend to provide stronger smoothing effects on optimization landscapes than residual connections alone. Hence, we might expect the additional skip connections present in NN also lead to a smoother optimization landscape.

In Figure~\ref{fig:opt_landscape_test} we illustrate the optimization landscapes using the method of \cite{li2018visualizing}, which proposes a procedure for projecting the loss landscape of very high-dimensional models into two dimensions for visualization purposes. Suppose that the fully-trained network has a set of parameters which is vectorized $\widehat{\btheta} \in \reals^K$. We draw two independent standard Gaussian vectors ${\bm v}_1$, ${\bm v}_2$ with dimension $K$, and normalize them in the manner described in \cite{li2018visualizing} that accounts for the scaling ambiguity of ReLU networks. Then we compute the test and training set error for parameters given by $\widehat{\btheta} + \tau (i {\bm v}_1 + j {\bm v}_2)$ for step size $\tau > 0$ and integers $i, j$. The plots above are generated for $i, j \in \{-125,..., 125\}$ and $\tau = 0.01$. We demonstrate plots for three different forward models: deblurring, compressed sensing with $8\times$ compression, and $10\times$ superresolution.

Figure~\ref{fig:opt_landscape_test} illustrates several attractive properties of the NN and GDN. Local minima appear to be rare in the neighborhood of the trained minima for GDN and NN. While neither is convex, it is interesting to note that the NN landscapes seem to have much wider basins around minima and higher slope outside this main basin. GDN's optimization landscape appears to require a search around a low-slope landscape until finding a region of high curvature in the deblurring and compressed sensing case, and contains more local minima than NN.

In addition, experimental evidence and some theory indicate that wider local minima have better generalization properties \cite{keskar2016large}. This does not indicate that one architecture should perform better than another, but if both networks achieve similar training error, wider local minima may translate to better test performance. 

\section{Discussion and Conclusions}
This paper describes a novel network architecture that departs from the currently-popular unrolled optimization framework described in Section~\ref{sec:unrolled}. Our approach is based on the Neumann series expansion for inverting linear operators and has several key features. First, Neumann networks naturally contain ``skip connections'' \cite{he2016deep,huang2017densely} that appear to yield optimization landscapes that facilitate more efficient training, but are absent from previously proposed network architectures.

Our theoretical analysis reveals that when the training data lie in a union of subspaces, the optimal {\em oracle} estimator that has prior knowledge of both the subspaces in the union and the identity of the subspace to which true image belongs is piecewise linear. We show this piecewise linear oracle estimator can be approximated arbitrarily well by a Neumann network whose learned component coincides with a specific piecewise linear map, which in principle can always be realized by a neural network using ReLU activations. Furthermore, we observe empirically on simulated union-of-subspaces data that the nonlinear learned component in the trained Neumann network well-approximates the specific piecewise linear map predicted by theory. 
We are unaware of past work on using neural networks to solve inverse problems demonstrating such properties.

Third, we describe a simple preconditioning step that, when combined with the Neumann network architecture, provides an additional increase in reconstruction PSNR and can reduce the number of blocks $B$ needed for accurate reconstruction, which in turn decreases reconstruction computational complexity when the preconditioning can be computed efficiently. As a result, using a truncated series expansion with only $B$ blocks results in a small, bounded approximation error. 
Finally, we explore the proposed Neumann network's empirical performance on a variety of inverse problems relative to the performance of representative agnostic, decoupled, and unrolled optimization methods described in Section~\ref{sec:previous}.

While this paper has focused on solving linear inverse problems in imaging using training data to train a neural network, {\em more generally we can think of this paper as a case study in leveraging physical models to guide neural network architecture design.} More specifically, we can think of networks such as the Neumann network as a single large neural network in which a subset of edge weights (\ie those corresponding to the operation $\bI-\eta\bX^\top\bX$ and other zero-valued ``edges" that define the general architecture of Figure~\ref{fig:nn}) are determined by the physical forward model that specifies the inverse problem at hand and are held fixed during training, while the remaining edges (\ie those that correspond to the operation $R(\cdot)$) can be learned during training. In other words, {\em we use knowledge of the inverse problem structure to define the neural network architecture.}

This perspective leads to interesting potential avenues for future work. Specifically, our proposed Neumann network is inspired by series expansions for inverting linear operators, but there are alternative methods for inverting nonlinear operators that may yield new challenges and opportunities. For instance, Adomian decompositions and polynomial expansions have been successfully used to solve differential equations with both linear and nonlinear components \cite{gabet1994theoretical,adomian2013solving}, and so designing future networks inspired by this framework could lead to new theoretical insights beyond what we present above. 

In addition, the reader might note that neural networks based on the Neumann series expansion or iterative optimization methods have several repeated blocks, leading to the question of whether standard stochastic gradient descent is the most efficient training regimen. For instance, recent work on ``Neural Ordinary Differential Equations'' \cite{neuralODE} has considered an ODE representation of the operation of a neural network instead of a series of discrete layers and used this perspective to devise training methods that leverage ODE solvers for more efficient training. Such techniques might be leveraged to improve training of Neumann networks.

\section*{Appendix}
Here we let $\bP_{\bX}$ and $\bP_{\bX_\perp}$ denote the projectors onto the row space of $\bX$ and the null space of $\bX$, respectively. In particular, $\bP_{\bX} = \bX^\T\bX$ and $\bP_{\bX_\perp} = \bI-\bX^\T\bX$, since we assume $\bX$ has orthonormal rows in Lemma 1, Theorem 1, and Corollary 1.
\subsection{Proof of Lemma 1}\label{sec:prooflem1}
We have $\bhat(\by) = \sum_{j=0}^B\btj{j}$ where $\btj{0}= \eta\bX^\T\by$, $\by = \bX\bstar$, and 
\begin{align}
    \btj{j} & = (\bI - \eta\bX^\T\bX -\eta R)\btj{j-1} \\
              & = (\bP_{\bX_\perp} + (1-\eta)\bP_{\bX} - \eta R)\btj{j-1}
\end{align}
for all $j=1,...,B$, and where in the last line we used the identity $\bI = \bP_{\bX} + \bP_{\bX_\perp}$.

We show that $R(\bbeta) = \bbR\bbeta$ with $\bbR$ as specified in Lemma 1 satisfies the desired error bounds. Define $\bQ = \bP_{\bX_\perp}\bU(\bU^\T\bX^\T\bX\bU)^{-1}\bU^\T\bX^\T$ so that $\bbR = -c_{\eta,B}\,\bQ\bX$. With this choice of $\bbR$ we have $\bP_{\bX} \bbR = 0$, and an easy induction shows
\begin{equation}\label{eq:betajs}
    \btj{j} = \eta(1-\eta)^j\bX^\T\by - \eta \sum_{k=0}^{j-1} \bbR\btj{k}
\end{equation}
for all $j\geq 1$. Summing this over $j=0,1,...,B$ gives
\begin{align}
    \bhat(\by) 
              & = \sum_{j=0}^B \eta(1-\eta)^j\bX^\T\by -\eta \sum_{j=1}^{B}\sum_{k=0}^{j-1} \bbR\btj{k}\nonumber\\
              & = \sum_{j=0}^B \eta(1-\eta)^j\bX^\T\by-\eta \sum_{j=0}^{B-1}(B-j)\bbR\btj{j}\label{eq:bhatlem}
\end{align}

Next, we show we can choose the constant $c_{\eta,B}$ so that the second term above simplifies to
\begin{equation}\label{eq:xperpeq}
   -\eta \sum_{j=0}^{B-1}(B-j)\bbR\btj{j} = \bQ\by.
\end{equation}
Observe that $\bbR^2\btj{k} = 0$ for all $k=0,...,j-1$, and so from \eqref{eq:betajs} we have $\bbR\btj{j} = \eta(1-\eta)^j\bbR\bX^\T\by$, which gives
\begin{equation*}
\sum_{j=0}^{B-1}(B-j)\bbR\btj{j} = \eta\sum_{j=0}^{B-1}(B-j)(1-\eta)^j\bbR\bX^\T\by.
\end{equation*}
Letting
$
c_{\eta,B} = \left(\eta^2\sum_{j=0}^{B-1}(B-j)(1-\eta)^j\right)^{-1}
$
we obtain \eqref{eq:xperpeq}. Therefore, combining \eqref{eq:bhatlem} and \eqref{eq:xperpeq} we have
\begin{align}\label{eq:finalform}
    \bhat(\by) & = \eta\sum_{j=0}^B (1-\eta)^j\bX^\T\by + \bQ\by.
\end{align}
Finally, since we assume $\by = \bX\bstar$, we see that $\bX^\T\by = \bP_{\bX}\bbeta^*$ and $\bQ\by = \bP_{\bX_\perp}\bbeta^*$, and using the fact that ${\eta\sum_{j=0}^B (1-\eta)^j = 1-(1-\eta)^{B+1}}$ from \eqref{eq:finalform} we have
$\bhat(\by) = \bstar - (1-\eta)^{B+1}\bP_{\bX}\bbeta^*$
which gives the desired error bound.

\subsection{Proof of Theorem \ref{thm:main} and Corollary \ref{cor:main}}

To prove Theorem~\ref{thm:main} we show that if $\bstar$ belongs to the $k$th subspace then $R^*$ acts acts as the linear map $\bbR_k$ when applied to each Neumann network  term $\btj{j}$. That is, we show  $\btj{j} \in \mathcal{C}_k$ for all $j=0,...,B$, where $\btj{0}= \eta\bX^\T\by$ and $\btj{j} = (\bI-\eta\bX^\T\bX)\btj{j-1} -\eta R^*(\btj{j-1})$ for $j=1,...,B$. The desired error bounds then follow by direct application of Lemma~\ref{lem:singlesub}.

First, an easy induction shows that
\begin{equation}
    \btj{j} = \eta(1-\eta)^j \bX^\T\by -\eta \sum_{i=0}^{j-1}R^*(\btj{k}).
\end{equation}
Using the fact that $\bP_{\bX}R^*(\bbeta) = 0$ for all $\bbeta \in \reals^p$, and $\by = \bX\bstar$, we have
$\bP_{\bX}\btj{j} = \eta(1-\eta)^j\bP_{\bX}\bstar$
for all $j=0,...,B$. Since the region $\mathcal{C}_k$ is a cone, in order to prove $\btj{j} \in \mathcal{C}_k$ for all $j=0,...,B$ it suffices to show $\bP_{\bX}\bstar \in \mathcal{C}_k$. This means we need to show $d_{\bX,k}(\bstar) < d_{\bX,\ell}(\bstar)$ for all $\ell\neq k$, or equivalently, 
\begin{multline*}
    \|(\bI-\bX\bU_k(\bX\bU_k)^{+})\bX\bstar\|\\ < \|(\bI-\bX\bU_\ell(\bX\bU_\ell)^{+})\bX\bstar\|
\end{multline*}
for all $\ell \neq k$. Since $\bX\bU_k(\bX\bU_k)^{+}$ is projection onto $\text{span}(\bX\bU_k)$, and $\bX\bstar \in \text{span}(\bX\bU_k)$, we have ${(\bI-\bX\bU_{k}(\bX\bU_{k})^{+})\bX\bstar = 0}$ and so
${\|(\bI-\bX\bU_{k}(\bX\bU_{k})^{+})\bX\bstar\| = 0}$.
Furthermore, since by assumption $\bX\bstar \notin \text{span}(\bX\bU_\ell)$ for all $\ell \neq k$, we have ${(\bI-\bX\bU_{k}(\bX\bU_{k})^{+})\bX\bstar \neq 0}$, which means
${\|(\bI-\bX\bU_\ell(\bX\bU_\ell)^{+})\bX\bstar\| > 0}$,
proving the claim.

Similarly, to prove Corollary~\ref{cor:main} we need to show $\bj{j} \in \mathcal{C}_k$ for all $j=0,...,B$, where $\bj{0} = \eta\bX^\T\by \in \mathcal{C}_k$ and $\bj{j} = (\bI-\eta\bX^\T\bX)\bj{j-1} -\eta R^*(\bj{j-1}) + \bj{0}$. Since $\bP_{\bX}R^*(\bbeta) = 0$ for all $\bbeta \in \reals^p$, an easy induction shows that 
\begin{equation}
    \bP_{\bX}\bj{j} = \eta\sum_{i=0}^j(1-\eta)^i\bP_{\bX}\bstar.
\end{equation}
Since each $\bj{j}$ is a scalar multiple of $\bP_{\bX}\bstar$, by the same argument as above we have $\bj{j} \in \mathcal{C}_k$ for all $j = 0,...,B$, which proves the claim.

\bibliographystyle{IEEEtran}
\bibliography{twocolumn_revision_black}

\section{Supplement}
\subsection{Additional Empirical Support for Union of Subspaces Theory}
In the experiment discussed in Sec.~IV, we train a Neumann network on pairs $(\bbeta_i,\by_i)_{i=1}^N$ where each $\bbeta_i\in \reals^{10}$ belongs one of three randomly chosen three-dimensional subspaces spanned by matrices $\bU_1,\bU_2,\bU_3\in\reals^{10\times 3}$ with orthonormal columns. We generate a random training point as $\bbeta_i = \bU_{k_i} \bw_i$ where we select the index $k_i \in \{1,2,3\}$ uniformly at random and $\bw_i \in \reals^3$ is a random vector with i.i.d. $\mathcal{N}(0,1)$ entries. Here we take $\bX \in \reals^{5\times 10}$ to be the first five rows of the $10\times 10$ identity matrix such that $\by_i = \bX\bbeta_i$ is the restriction of $\bbeta_i$ to its first five coordinates. We train a $6$-block ($B=6$) Neumann network  where the learned component $R:\reals^{10}\rightarrow\reals^{10}$ is a seven-layer fully-connected neural network with ReLU activations such that the five hidden layers have sizes $(10,10,6,10,10)$. We learn a set of weights for the learned component and the Neumann network step size $\eta$ by minimizing the empirical risk using SGD with ADAM acceleration and a batch size of 64, training for 100,000 epochs. When evaluated on a test set of $M=1024$ points drawn at random from the union of subspaces in the same manner as the training set, the learned component achieves mean squared error $\frac{1}{M}\sum_{i=1}^M \|\bhat(\by_i)-\bbeta_i\|^2 =  0.0176$ with variance 0.001, indicating the trained network learned to accurately reconstruct inputs belonging to the union of subspaces.

\begin{figure*}[ht!]
\centering
\begin{tabular}[t]{ccc}
\includegraphics[height=0.32\textwidth]{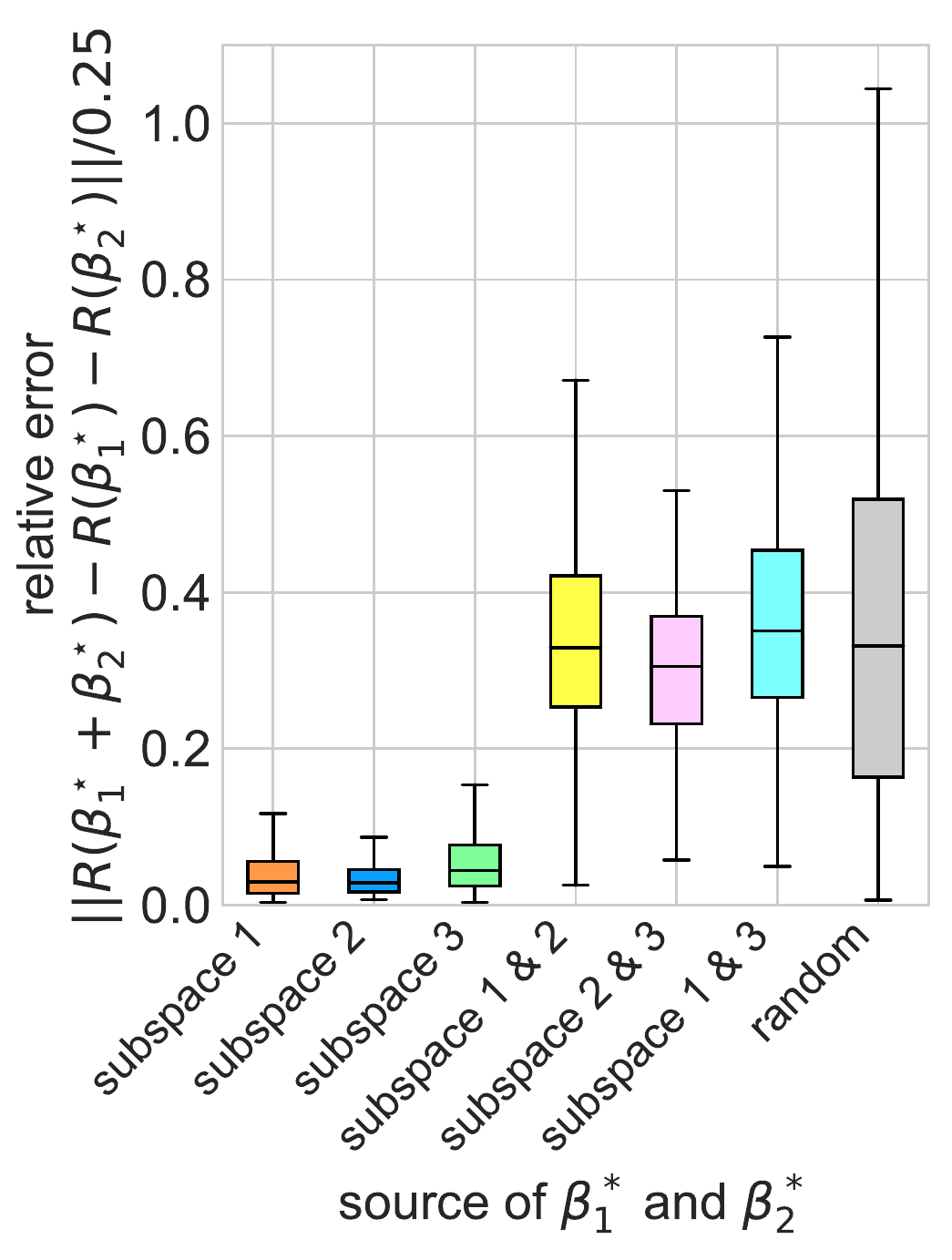} & 
\raisebox{1em}{\includegraphics[height=0.28\textwidth]{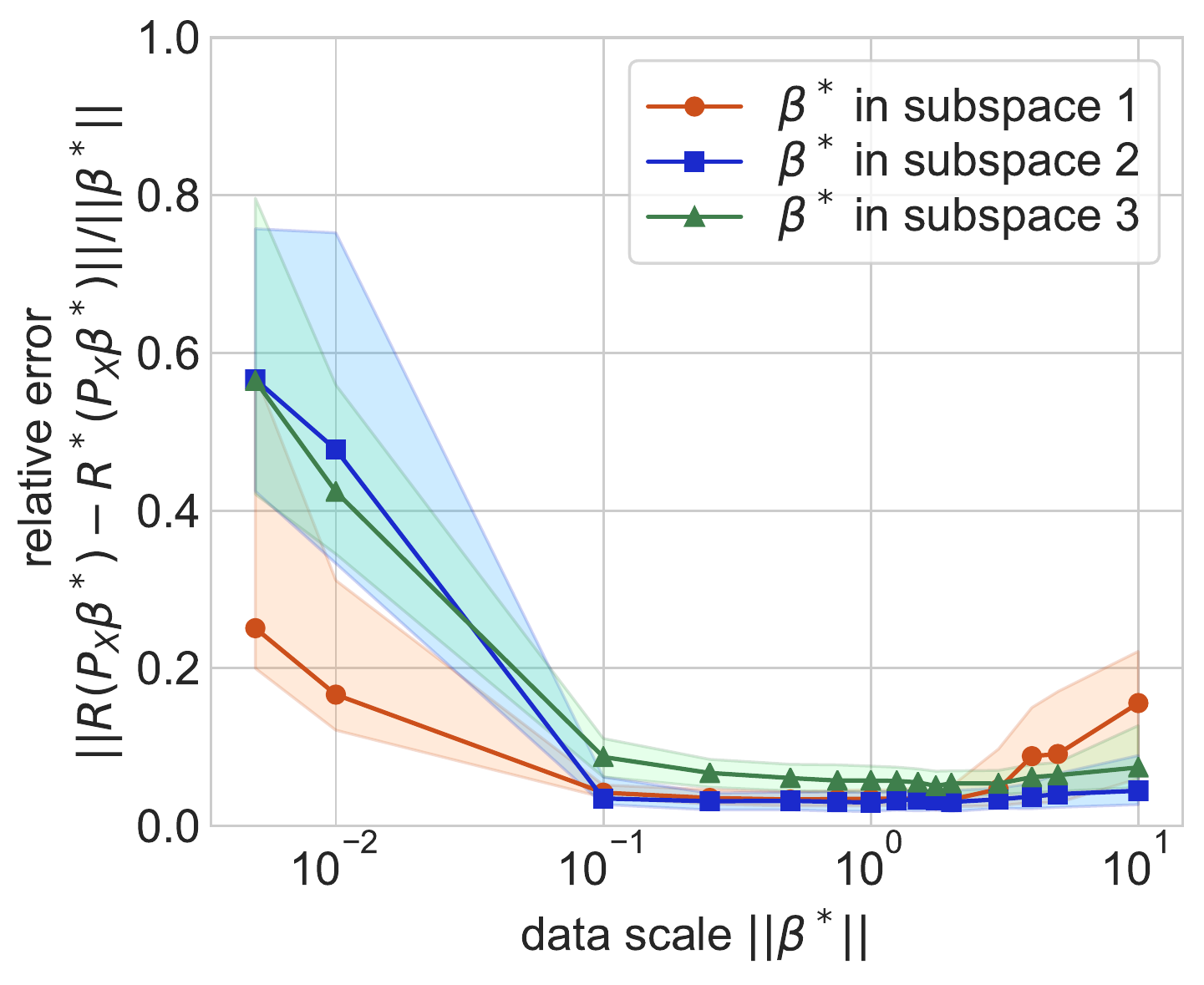}} & 
\raisebox{1em}{\includegraphics[height=0.28\textwidth]{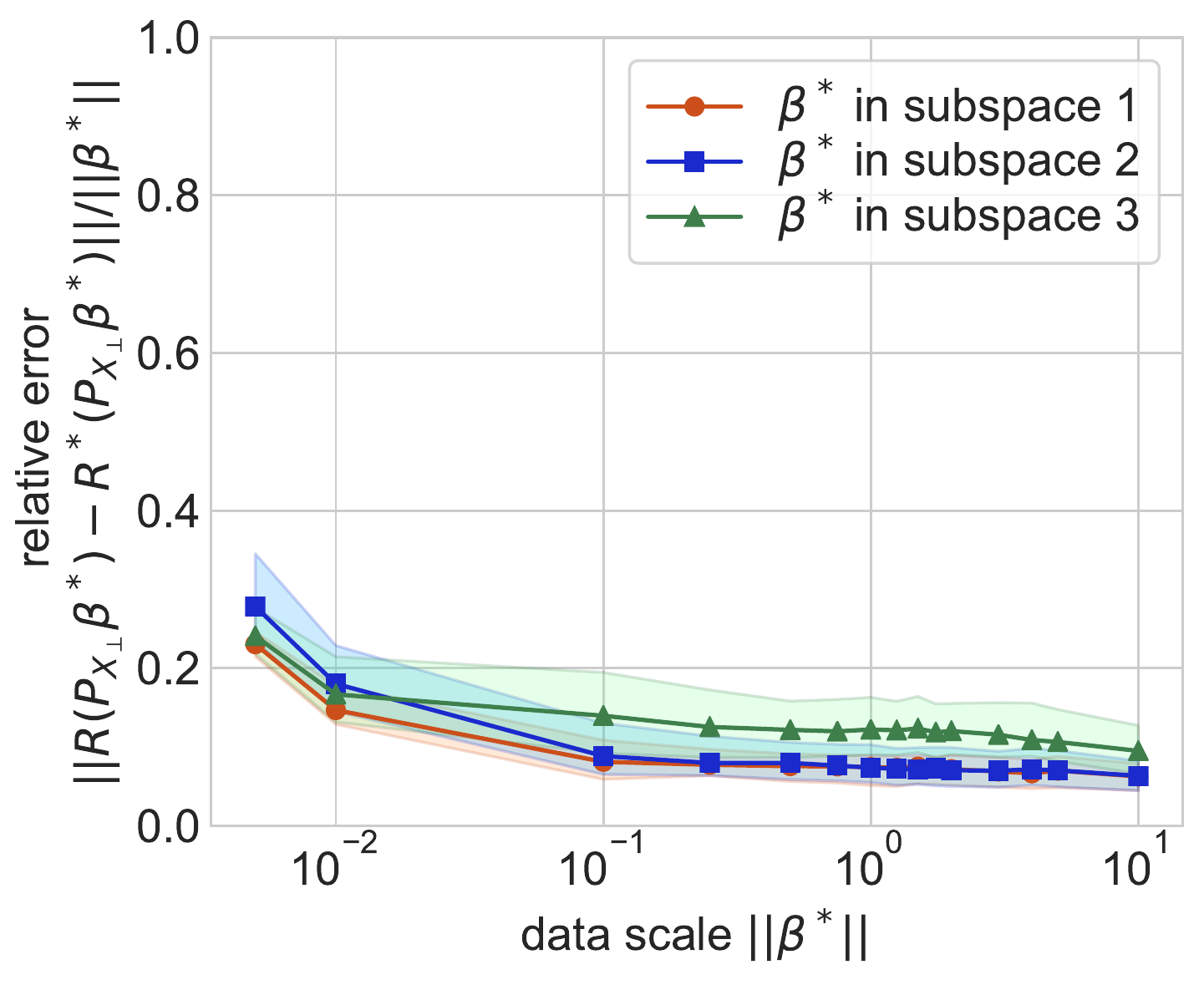}}\\ 
(a) Piecewise linearity test & (b) Input in row space of $\bX$  & (c) Input in null space of $\bX$
\end{tabular}
\caption{\small {Empirical support for Theorem 1}. We train a Neumann network   on synthetic ``images'' (vectors in $\reals^{10}$) belonging to a union of subspaces for a 1-D inpainting problem ($K=3$ random subspaces each of dimension $r=3$, $m=5$ measurements). Our theory predicts that the learned component $R$ should be close to the piecewise linear mapping $R^*$ defined in Theorem~1, and we perform three empirical tests to see if this is true. In (a) we measure how linear $R$ is when evaluated at two vectors drawn from the same subspace, from two different subspaces, or from two random Gaussian vectors. The plot illustrates that the learned $R$ only behaves like a linear operator when the vectors belong the same subspace (\ie the relative error is small), which indicates $R$ is approximately piecewise linear. In (b) we demonstrate that the learned $R$ behaves like $R^*$ when restricted to inputs of the form $\bP_{\bX}\bstar$ (projection of $\bstar$ onto the row space of $\bX$), over several $\bstar$ drawn from each subspace uniformly at random, normalized to different scales $\|\bstar\|$. Similarly, in (c) we demonstrate the output of the learned $R$ is close to the output of $R^*$ when restricted to inputs of the form $\bP_{\bX_\perp}\bstar$ (projection of $\bstar$ onto the null space of $\bX$) over a range of scales. In (a) is a box-and-whisker plot of the results from 1024 random trials for each input source. The points plotted in (b) and (c) are the median of the relative error computed over 1024 random trials at that scale, with the shaded region indicating the interquartile range. The learned component was trained on vectors with norm approximately $1$, yet we find the learned $R$ generalizes across a range of scales.}
\label{fig:learnedR_exp1_supp}
\end{figure*}

Figure~\ref{fig:learnedR_exp1_supp} displays the results of three quantitative experiments to assess whether the learned component $R$ behaves as the piecewise linear $R^*$ predicted by Theorem 1.
First, we test whether the learned $R$ is approximately linear when restricted to inputs belonging to each subspace, \ie we test whether $R(\bstar_1 + \bstar_2) \approx R(\bstar_1) + R(\bstar_2)$, for all $\bstar_1,\bstar_2$ belonging to the same subspace. As baselines we compare to the case where $\bstar_1$ and $\bstar_2$ belong to different subspaces, and the case where $\bstar_1$ and $\bstar_2$ are Gaussian random vectors. In Figure~\ref{fig:learnedR_exp1_supp}(a) we display a boxplot of the relative error $\|R(\bstar_1 + \bstar_2) -R(\bstar_1)-R(\bstar_2)\|/\gamma$ of 1024 randomly generated $\bstar_1,\bstar_2$ normalized such that $\|\bstar_1\|=\|\bstar_2\| = \gamma$ for the various cases. Here we set normalization to $\gamma = 0.25$, though similar results were obtained for $\gamma \in [0.1,0.5]$ (not shown). As predicted, the relative error concentrates near zero in the case where $\bstar_1,\bstar_2$ belong to the same subspace, and is otherwise large, indicating the learned $R$ is indeed approximately piecewise linear as predicted by Theorem 1.

The $R^*$ specified in Theorem 1 acts differently on vectors in the row space of $\bX$ and the nullspace of $\bX$. We perform two experiments to verify this is true of our learned $R$ as well.

First, we evaluate $R$ on inputs of the form $\bP_{\bX}\bstar$ where $\bstar$ belongs to one of the three subspaces and $\bP_{\bX}:=\bX^\T\bX$ is the projection onto the row space of $\bX$. If $\bstar$ belongs to one of the three subspaces we have $R^*(\bP_{\bX}\bstar) = -c_{\eta,B}\bP_{\bX_\perp}\bstar$, where $c_{\eta,B}$ is a constant depending on $\eta$ and $B$. In the present setting ($B=6$ blocks, and learned $\eta = 0.482$) we have $c_{\eta,B} = 0.349$. In Figure~\ref{fig:learnedR_exp1_supp}(b), we plot the median of the relative error $\|R(\bP_{\bX}\bstar) - R^*(\bP_{\bX}\bstar)\|/\|\bstar\|$ where $\bstar$ is normalized to different scales $10^{-3} \leq \|\bstar\| \leq 10$. Observe that the relative error is small over a wide range of scales, even though the network was trained on inputs $\bstar$ with $\|\bstar\|\approx 1$, which indicates the learned $R$ generalizes well to other scales.

Next, we evaluate $R$ on inputs of the form $\bP_{\bX_\perp}\bstar$ where $\bP_{\bX_\perp} = \bI-\bX^\T\bX$ denotes projection onto the nullspace of $\bX$. The predicted output in this case is $R^*(\bP_{\bX_\perp}\bstar) = 0$. In Figure~\ref{fig:learnedR_exp1_supp}(c), we plot the median of the relative error $\|R(\bP_{\bX_\perp}\bstar)-R^*(\bP_{\bX_\perp}\bstar)|/\|\bstar\|$ of 1024 randomly generated $\bstar$ normalized to various scales. In this case, we find the relative error is low over all scales, again indicating good generalization of the learned $R$.

\subsection{Backpropagation Gradients of Neumann and Gradient Descent Networks}\label{sec:backprop}
Let $\bhat(\by;\btheta)$ be a Neumann network estimator depending on parameters $\btheta$ (\ie the parameters defining regularizer network $R(\bbeta;\btheta)$). 
Gradients of the empirical risk $\mathcal{L}(\btheta) = {\sum_{i=1}^N \|\bhat(\by_i;\btheta)-\bbeta_i\|}$ have the form
\begin{equation}
    \frac{\partial \mathcal{L}(\btheta)}{\partial \btheta} = \sum_{i=1}^N\frac{\partial \bhat(\by_i; \btheta)}{\partial \btheta}^\T \left(\bhat(\by_i;\btheta)-\bbeta_i\right)
\end{equation}
where $\frac{\partial \bhat(\by_i; \btheta)}{\partial \btheta}$ is the Jacobian of $\bhat$ with respect to $\btheta$ evaluated at $(\by_i;\btheta)$.

Dropping the dependence on $\by_i$, we write the Neumann network  as a sum $\bhat(\btheta) = \sum_{j=0}^B \btj{j}(\btheta)$ where 
\begin{equation}\label{eq:recursion}
\btj{j+1}(\btheta) = (\bI - \eta\bX^\T\bX)\btj{j}(\btheta) - \eta R(\btj{j}(\btheta);\btheta)
\end{equation}
with $\btj{0} = \eta\bX^\T\by_i$ and where $R(\btheta;\bbeta)$ denotes the learned component depending on parameters $\btheta$ evaluated at input $\bbeta$. Hence, we have
\begin{equation}
    \frac{\partial \bhat(\by; \btheta)}{\partial \btheta} = \sum_{j=0}^B \frac{\partial \btj{j}(\btheta)}{\partial \btheta}
\end{equation}
The zero order term vanishes because it is constant (assuming $\eta$ is fixed). To compute the Jacobian of the $\btj{j}$, $j\geq 1$, we use the recursive formula \eqref{eq:recursion} and the chain rule to get:
\begin{align}
    \frac{\partial \btj{j+1}(\btheta)}{\partial \btheta} & = (\bI - \eta\bX^\T\bX)\frac{\partial \btj{j}(\btheta)}{\partial \btheta} - \eta \left(\frac{\partial R(\btj{j}(\btheta);\btheta)}{\partial \btheta} + \frac{\partial R(\btj{j}(\btheta);\btheta)}{\partial \bbeta}\frac{\partial \btj{j}(\btheta)}{\partial \btheta}\right)\\
    & = \left(\bI - \eta\bX^\T\bX - \eta \frac{\partial R(\btj{j}(\btheta);\btheta)}{\partial \bbeta}\right)\frac{\partial \btj{j}(\btheta)}{\partial \btheta} - \eta \frac{\partial R(\btj{j}(\btheta);\btheta)}{\partial \btheta}
\end{align}
To simplify notation, define $F(\bbeta) = (\bI - \eta\bX^T\bX - \eta R)(\bbeta)$, suppressing the dependence of $R$ on $\btheta$, and we write $\partial_{\bbeta} F|_{\bbeta=\bstar}$ for the Jacobian of $F$ with respect to $\bbeta$ evaluated at $\bstar$ with the current value of $\btheta$. Similarly, write $\partial_{\btheta} R|_{\bbeta = \bstar}$ for the Jacobian of $R$ with respect to $\btheta$ evaluated at $\bbeta = \bstar$ with the current value of $\btheta$. In this notation, the above becomes:
\begin{equation}
    \partial_{\btheta}\btj{j+1} = \partial_{\bbeta} F |_{\bbeta = \btj{j}} \cdot \partial_{\btheta}\btj{j} - \eta\, \partial_{\btheta} R|_{\bbeta = \btj{j}}
\end{equation}
For example,
\begin{equation}
    \partial_{\btheta}\btj{1} = - \eta\, \partial_{\btheta} R|_{\bbeta = \btj{0}}
\end{equation}
and
\begin{equation}
    \partial_{\btheta}\btj{2} = - \eta\left(\partial_{\bbeta} F |_{\bbeta = \btj{1}} \cdot \partial_{\btheta} R|_{\bbeta = \btj{0}} + \partial_{\btheta} R|_{\bbeta = \btj{1}}\right)
\end{equation}
and 
\begin{equation}
    \partial_{\btheta}\bbeta_{3} = -\eta\left(\partial_{\bbeta} F |_{\bbeta = \btj{2}} \cdot \partial_{\bbeta} F |_{\bbeta = \btj{1}} \cdot \partial_{\btheta} R|_{\bbeta = \btj{0}} + \partial_{\bbeta} F |_{\bbeta = \btj{2}}\cdot \partial_{\btheta} R|_{\bbeta = \btj{1}}+ \partial_{\btheta} R|_{\bbeta = \btj{2}}\right)
\end{equation}
and so in general
\begin{equation}
    \partial_{\btheta}\bbeta_{k} = -\eta \left(\sum_{j'=0}^{j-1} \left(\partial_{\bbeta}F |_{\bbeta = \btj{j-1}}\cdot \partial_{\bbeta}F |_{\bbeta = \btj{j-2}}\cdot \ \cdots \ \cdot \partial_{\bbeta}F |_{\bbeta = \btj{j'+1}}\right)\partial_{\btheta} R|_{\bbeta = \btj{j'}} \right).
\end{equation}
Therefore,
\begin{equation}
    \partial_{\btheta}\bhat = -\eta \sum_{j=0}^{B} \left(\sum_{j'=0}^{j-1} \left(\partial_{\bbeta}F |_{\bbeta = \btj{j-1}}\cdot \partial_{\bbeta}F |_{\bbeta = \btj{j-2}}\cdot \ \cdots \ \cdot \partial_{\bbeta}F |_{\bbeta = \btj{j'+1}}\right)\partial_{\btheta} R|_{\bbeta = \btj{j'}} \right).
\end{equation}

Now we perform the same analysis for the gradient descent network. We write a $B$-block gradient descent network as $\bhat'(\btheta) = \bj{B}(\btheta)$ where 
\begin{equation}\label{eq:recursion_gd}
\bj{j+1}(\btheta) = (\bI - \eta\bX^\T\bX)\bj{j}(\btheta) - \eta R(\bj{j}(\btheta); \btheta) + \bj{0}
\end{equation}
with $\bj{0} = \eta\bX^\T\by_i$ and where $R(\bbeta; \btheta)$ denotes the learned component depending on parameters $\btheta$ evaluated at input $\bbeta$. Similar to the Neumann network case, the derivatives have the recursive formula:
\begin{equation}
    \partial_{\btheta}\bj{j+1} = \partial_{\bbeta} F |_{\bbeta = \bj{j}} \cdot \partial_{\btheta}\bj{j} - \eta\, \partial_{\btheta} R|_{\bbeta = \bj{j}}
\end{equation}
And so
\begin{equation}
    \partial_{\btheta}\bhat' = -\eta \left(\sum_{j'=0}^{B-1} \left(\partial_{\bbeta}F |_{\bbeta = \bj{j-1}}\cdot \partial_{\bbeta}F |_{\bbeta = \bj{j-2}}\cdot \ \cdots \ \cdot \partial_{\bbeta}F |_{\bbeta = \bj{j'+1}}\right)\partial_{\btheta} R|_{\bbeta = \bj{j'}} \right).
\end{equation}

To compare the two gradient updates, for a $B = 3$ block Neumann network  we have
\begin{multline}
    \partial_{\btheta}\bhat =  -\eta\bigg(\partial_{\btheta}R|_{\bbeta = \btj{0}} + \partial_{\btheta}R|_{\bbeta = \btj{1}} + \partial_{\btheta}R|_{\bbeta = \btj{2}}\\
     + \partial_{\bbeta} F |_{\bbeta = \btj{1}} \cdot \partial_{\btheta} R|_{\bbeta = \btj{0}} + \partial_{\bbeta} F |_{\bbeta = \bbeta_2} \cdot \partial_{\btheta} R|_{\bbeta = \btj{1}}\\
     + \partial_{\bbeta} F |_{\bbeta = \btj{2}} \cdot \partial_{\bbeta} F |_{\bbeta = \btj{1}} \cdot \partial_{\btheta} R|_{\bbeta = \btj{0}}\bigg).
\end{multline}
and for $B = 3$ block gradient descent network we have
\begin{equation}
    \partial_{\btheta}\bhat' = -\eta\left(\partial_{\btheta} R|_{\bbeta = \bj{2}} +  \partial_{\bbeta} F |_{\bbeta = \bj{2}}\cdot \partial_{\btheta} R|_{\bbeta = \bj{1}}+ \partial_{\bbeta} F |_{\bbeta = \bj{2}} \cdot \partial_{\bbeta} F |_{\bbeta = \bj{1}} \cdot \partial_{\btheta} R|_{\bbeta = \bj{0}}\right)
\end{equation}

\section{Implementation Details of the Learned Component of the Neumann and Gradient Descent Networks}

Implementation details for the learned components of the Neumann and gradient descent networks are described here, along with details of the ResAuto architecture.

For all networks, bias initialization was a constant at $0.001$, while other weights were initialized with a truncated normal with variance 0.05 for Neumann networks and ResAuto, and 0.01 for gradient descent networks. Weight decay was not used.

The learned components of all networks used were convolutional-deconvolutional networks, similar to those used in \cite{chen2017low, chang2017one}. The network structures can be described by the sizes and strides of the filters used, along with the nonlinearities. The nonlinearity was chosen to be the ELU \cite{clevert2015fast}.

\sloppy The filters on the network used for CIFAR-10 are square with sizes $5, 3, 2, 3,3,5,3$ and strides $1,2,1,2,1,1,1$. The first three layers convolutional layers, and the last four are convolution transposed (``deconvolution") layers. The number of outputs of each layer are $64, 256, 256, 256, 256, 64, 3$. The channelwise fully-connected layer is after the third, final convolutional layer. The filters on the network used for STL-10 and CelebA are similar to the architecture used on CIFAR-10, except for some adjustments to the number of outputs at each layer, which are $64, 256, 512, 512, 256, 64, 3$.

In the Residual Autoencoder, the filters were square with sizes $4,4,4,4,4,2,4,4,4,4,4$, with strides $1,1,2,2,2,1,1,2,2,2,1$ and output sizes $32, 64, 128, 256, 512, 512, 512, 256, 128, 64, 3$. There are residual connections between the second layer and the penultimate layer, and between the input and output layers.

Residual connections were not used in the learned component inside the Neumann network because each block of the network contains an implicit residual connection. Specifically, each block contains an operator of the form $-\eta \bX^\top \bX + (\bI-\eta R)$, where the $R$ is the learned component. Aside from the scaling by $-\eta$, the second term is identical to a network with a residual connection from the input to the output. For almost identical reasons, including a residual connection in the gradient descent network learned component would be redundant as well.

Training was done in Tensorflow, and optimized by ADAM with a beginning step size that was tuned for each problem and decayed exponentially with a decay rate of 0.99 per 500 steps. Beginning step sizes for Neumann networks and gradient descent networks tended to be around $10^{-3}$, while the step size for the Residual Autoencoder was closer to $10^{-1}$. Anecdotally, gradient descent networks tended to be more sensitive to initialization and step size tuning.

Training was run for 50 epochs across all datasets. Batch size was 32 for the CIFAR tests, and 16 for CelebA and STL networks, due to memory constraints. All training was done on Amazon EC2 p2.xlarge instances on a single NVidia Tesla K80 GPU with 12 GB of GPU memory.

\subsection{Time Requirements Comparison of Inverse Problem Solvers}

 We present here a comparison of the time required to train and test several inverse problem solution methods. The results presented here should be considered relative only to each other, and may be different from problem to problem because of the difference in runtimes for forward problems.

\begin{table}[htbp]
\centering
\begin{tabular}{l |llllll}
 & PNN & NN & GDN & TNRD & MoDL & ResAuto  \\ \hline
Train (hours)  & 4.6 & 4.2 & 3.9 & 1.4 & 4.7 & 1.1  \\
Test (sec) & 7.0 & 5.9 & 5.8 & 2.2 & 6.6 & 1.4 \\ \hline
\end{tabular}\vspace{1em}
\caption{\small Train and testing times for the deblur plus noise reconstruction problem on the CIFAR10 dataset. The train times are in hours, while the test times are listed in seconds for a single batch of size 32.}
\label{table:timing_problem}
\end{table}

The time required to train and test any one of the unrolled methods presented here is dependent on a variety of factors. Briefly, one may consider the number of blocks $B$, the time to run the forward gramian operator $X^\top X$, the complexity of the learned component, and the computational resources at hand. For example, if GPU time is freely available but the forward model is not easily parallelizable and expensive to run, the dominant factor in reconstruction may be the data-consistency terms in all iterative methods.

Overall, however, we find that results are largely intuitive: preconditioning requires more time and for our settings, the learned components dominate time requirements for training. TNRD and ResAuto have approximately the same number of trainable parameters, but TNRD's additional linear components add time.

\subsection{Qualitative Comparison of Inverse Problem Solvers}

\begin{figure}[h]
\centering
\renewcommand{\arraystretch}{0.2}
\begin{tabular}{c|@{}c@{}c@{}c@{}c@{}c@{}c}
 & {\small Inpaint} & {\small Deblur} & {\small CS2} & {\small CS8} & {\small SR4} & {\small SR10} \\ \hline
  \raisebox{1.25em}{Original} & \subfloat{\includegraphics[width = 0.07\textwidth]{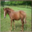}} &
\subfloat{\includegraphics[width = 0.07\textwidth]{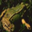}} &
\subfloat{\includegraphics[width = 0.07\textwidth]{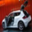}} &
\subfloat{\includegraphics[width = 0.07\textwidth]{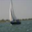}} &
\subfloat{\includegraphics[width = 0.07\textwidth]{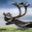}} &
\subfloat{\includegraphics[width = 0.07\textwidth]{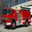}}\\[-1em]
\raisebox{1.25em}{$\bX^\top \by$} & \subfloat{\includegraphics[width = 0.07\textwidth]{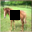}} &
\subfloat{\includegraphics[width = 0.07\textwidth]{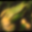}} &
\subfloat{\includegraphics[width = 0.07\textwidth]{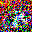}} &
\subfloat{\includegraphics[width = 0.07\textwidth]{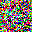}} &
\subfloat{\includegraphics[width = 0.07\textwidth]{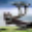}} &
\subfloat{\includegraphics[width = 0.07\textwidth]{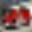}}\\[-1em]
\raisebox{1.25em}{NN} & \subfloat{\includegraphics[width = 0.07\textwidth]{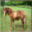}} &
\subfloat{\includegraphics[width = 0.07\textwidth]{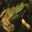}} &
\subfloat{\includegraphics[width = 0.07\textwidth]{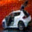}} &
\subfloat{\includegraphics[width = 0.07\textwidth]{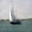}} &
\subfloat{\includegraphics[width = 0.07\textwidth]{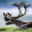}} &
\subfloat{\includegraphics[width = 0.07\textwidth]{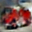}}\\[-1em]
\raisebox{1.25em}{GDN} & \subfloat{\includegraphics[width = 0.07\textwidth]{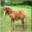}} &
\subfloat{\includegraphics[width = 0.07\textwidth]{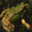}} &
\subfloat{\includegraphics[width = 0.07\textwidth]{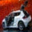}} &
\subfloat{\includegraphics[width = 0.07\textwidth]{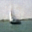}} &
\subfloat{\includegraphics[width = 0.07\textwidth]{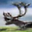}} &
\subfloat{\includegraphics[width = 0.07\textwidth]{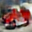}}\\[-1em]
\raisebox{1.25em}{TNRD} & \subfloat{\includegraphics[width = 0.07\textwidth]{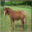}} &
\subfloat{\includegraphics[width = 0.07\textwidth]{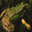}} &
\subfloat{\includegraphics[width = 0.07\textwidth]{figures/pretty_pictures/cifar/cs2/grad}} &
\subfloat{\includegraphics[width = 0.07\textwidth]{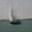}} &
\subfloat{\includegraphics[width = 0.07\textwidth]{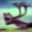}} &
\subfloat{\includegraphics[width = 0.07\textwidth]{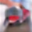}}\\[-1em]
\raisebox{1.25em}{MoDL} & \subfloat{\includegraphics[width = 0.07\textwidth]{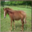}} &
\subfloat{\includegraphics[width = 0.07\textwidth]{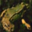}} &
\subfloat{\includegraphics[width = 0.07\textwidth]{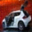}} &
\subfloat{\includegraphics[width = 0.07\textwidth]{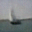}} &
\subfloat{\includegraphics[width = 0.07\textwidth]{figures/pretty_pictures/cifar/sample2/grad}} &
\subfloat{\includegraphics[width = 0.07\textwidth]{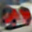}}\\[-1em]
\raisebox{1.25em}{ResAuto} & \subfloat{\includegraphics[width = 0.07\textwidth]{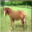}} &
\subfloat{\includegraphics[width = 0.07\textwidth]{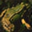}} &
\subfloat{\includegraphics[width = 0.07\textwidth]{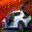}} &
\subfloat{\includegraphics[width = 0.07\textwidth]{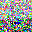}} &
\subfloat{\includegraphics[width = 0.07\textwidth]{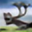}} &
\subfloat{\includegraphics[width = 0.07\textwidth]{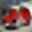}}\\[-1em]
\raisebox{1.25em}{CSGM} & \subfloat{\includegraphics[width = 0.07\textwidth]{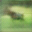}} &
\subfloat{\includegraphics[width = 0.07\textwidth]{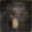}} &
\subfloat{\includegraphics[width = 0.07\textwidth]{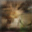}} &
\subfloat{\includegraphics[width = 0.07\textwidth]{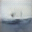}} &
\subfloat{\includegraphics[width = 0.07\textwidth]{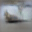}} &
\subfloat{\includegraphics[width = 0.07\textwidth]{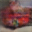}}\\[-1em]
\raisebox{1.25em}{TV} & \subfloat{\includegraphics[width = 0.07\textwidth]{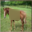}} &
\subfloat{\includegraphics[width = 0.07\textwidth]{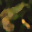}} &
\subfloat{\includegraphics[width = 0.07\textwidth]{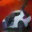}} &
\subfloat{\includegraphics[width = 0.07\textwidth]{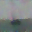}} &
\subfloat{\includegraphics[width = 0.07\textwidth]{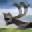}} &
\subfloat{\includegraphics[width = 0.07\textwidth]{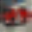}}\\
\end{tabular}
\caption{\small Visual demonstration of inverse problem solutions on the CIFAR-10 dataset. The Network Input row represents $\bX^\top \by$, which is fed into GDN, NN, ResAuto, and TNRD. Inpainting has a $10 \times 10$ inpainting region. Deblur has a Gaussian convolutional filter with $\sigma = 2.5$ and filter dimensions $5\times 5$. CS2 has a compression ratio of 2 with a Gaussian sensing matrix, CS8 has a compression ratio of 8 also with a Gaussian sensing matrix. SR (4x) downsamples by a factor of 2 along both dimensions using a linear filter. SR (10x) downsamples by a factor of $\sqrt{10}$ along both dimensions and also uses a linear filter.}
\label{fig:cifar_qual}
\end{figure}

\begin{figure}
\centering
\renewcommand{\arraystretch}{0.5}
\begin{tabular}{c|c@{}c@{}c@{}c@{}c@{}c@{}c@{}c@{}c@{}c@{}c@{}c}
 & {\small Inpaint} & {\small Deblur} & {\small CS2} & {\small CS8} & {\small SR4} & {\small SR10} & {\small Inpaint} & {\small Deblur} & {\small CS2} & {\small CS8} & {\small SR4} & {\small SR10} \\ \hline
  \raisebox{1.25em}{Original} & \subfloat{\includegraphics[width = 0.07\textwidth]{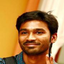}} &
\subfloat{\includegraphics[width = 0.07\textwidth]{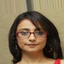}} &
\subfloat{\includegraphics[width = 0.07\textwidth]{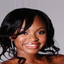}} &
\subfloat{\includegraphics[width = 0.07\textwidth]{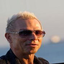}} &
\subfloat{\includegraphics[width = 0.07\textwidth]{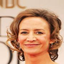}} &
\subfloat{\includegraphics[width = 0.07\textwidth]{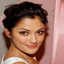}} &
\subfloat{\includegraphics[width = 0.07\textwidth]{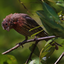}} &
\subfloat{\includegraphics[width = 0.07\textwidth]{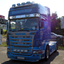}} &
\subfloat{\includegraphics[width = 0.07\textwidth]{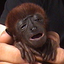}} &
\subfloat{\includegraphics[width = 0.07\textwidth]{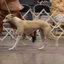}} &
\subfloat{\includegraphics[width = 0.07\textwidth]{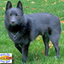}} &
\subfloat{\includegraphics[width = 0.07\textwidth]{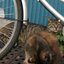}}\\[-1em]
\raisebox{1.25em}{$\bX^\top \by$} & \subfloat{\includegraphics[width = 0.07\textwidth]{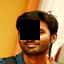}} &
\subfloat{\includegraphics[width = 0.07\textwidth]{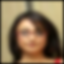}} &
\subfloat{\includegraphics[width = 0.07\textwidth]{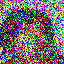}} &
\subfloat{\includegraphics[width = 0.07\textwidth]{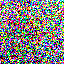}} &
\subfloat{\includegraphics[width = 0.07\textwidth]{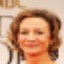}} &
\subfloat{\includegraphics[width = 0.07\textwidth]{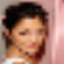}} &
\subfloat{\includegraphics[width = 0.07\textwidth]{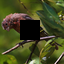}} &
\subfloat{\includegraphics[width = 0.07\textwidth]{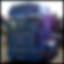}} &
\subfloat{\includegraphics[width = 0.07\textwidth]{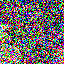}} &
\subfloat{\includegraphics[width = 0.07\textwidth]{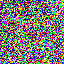}} &
\subfloat{\includegraphics[width = 0.07\textwidth]{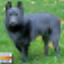}} &
\subfloat{\includegraphics[width = 0.07\textwidth]{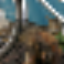}}\\[-1em]
\raisebox{1.25em}{NN} & \subfloat{\includegraphics[width = 0.07\textwidth]{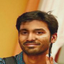}} &
\subfloat{\includegraphics[width = 0.07\textwidth]{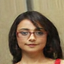}} &
\subfloat{\includegraphics[width = 0.07\textwidth]{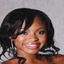}} &
\subfloat{\includegraphics[width = 0.07\textwidth]{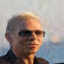}} &
\subfloat{\includegraphics[width = 0.07\textwidth]{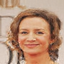}} &
\subfloat{\includegraphics[width = 0.07\textwidth]{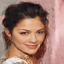}} &
\subfloat{\includegraphics[width = 0.07\textwidth]{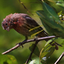}} &
\subfloat{\includegraphics[width = 0.07\textwidth]{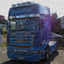}} &
\subfloat{\includegraphics[width = 0.07\textwidth]{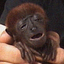}} &
\subfloat{\includegraphics[width = 0.07\textwidth]{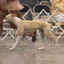}} &
\subfloat{\includegraphics[width = 0.07\textwidth]{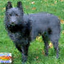}} &
\subfloat{\includegraphics[width = 0.07\textwidth]{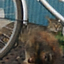}}\\[-1em]
\raisebox{1.25em}{GDN} & \subfloat{\includegraphics[width = 0.07\textwidth]{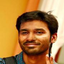}} &
\subfloat{\includegraphics[width = 0.07\textwidth]{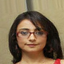}} &
\subfloat{\includegraphics[width = 0.07\textwidth]{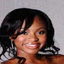}} &
\subfloat{\includegraphics[width = 0.07\textwidth]{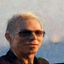}} &
\subfloat{\includegraphics[width = 0.07\textwidth]{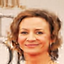}} &
\subfloat{\includegraphics[width = 0.07\textwidth]{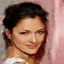}} &
\subfloat{\includegraphics[width = 0.07\textwidth]{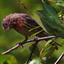}} &
\subfloat{\includegraphics[width = 0.07\textwidth]{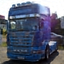}} &
\subfloat{\includegraphics[width = 0.07\textwidth]{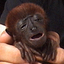}} &
\subfloat{\includegraphics[width = 0.07\textwidth]{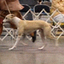}} &
\subfloat{\includegraphics[width = 0.07\textwidth]{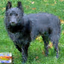}} &
\subfloat{\includegraphics[width = 0.07\textwidth]{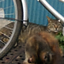}}\\[-1em]
\raisebox{1.25em}{ResAuto} & \subfloat{\includegraphics[width = 0.07\textwidth]{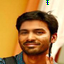}} &
\subfloat{\includegraphics[width = 0.07\textwidth]{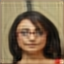}} &
\subfloat{\includegraphics[width = 0.07\textwidth]{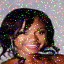}} &
\subfloat{\includegraphics[width = 0.07\textwidth]{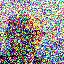}} &
\subfloat{\includegraphics[width = 0.07\textwidth]{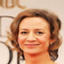}} & 
\subfloat{\includegraphics[width = 0.07\textwidth]{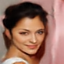}} &
\subfloat{\includegraphics[width = 0.07\textwidth]{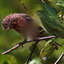}} &
\subfloat{\includegraphics[width = 0.07\textwidth]{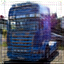}} &
\subfloat{\includegraphics[width = 0.07\textwidth]{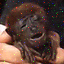}} &
\subfloat{\includegraphics[width = 0.07\textwidth]{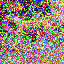}} &
\subfloat{\includegraphics[width = 0.07\textwidth]{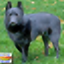}} &
\subfloat{\includegraphics[width = 0.07\textwidth]{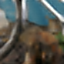}}\\[-1em]
\raisebox{1.25em}{CSGM} & \subfloat{\includegraphics[width = 0.07\textwidth]{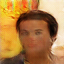}} & 
\subfloat{\includegraphics[width = 0.07\textwidth]{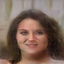}} &
\subfloat{\includegraphics[width = 0.07\textwidth]{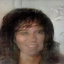}} &
\subfloat{\includegraphics[width = 0.07\textwidth]{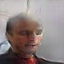}} &
\subfloat{\includegraphics[width = 0.07\textwidth]{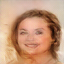}} &
\subfloat{\includegraphics[width = 0.07\textwidth]{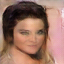}} &
\subfloat{\includegraphics[width = 0.07\textwidth]{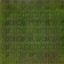}} &
\subfloat{\includegraphics[width = 0.07\textwidth]{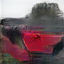}} &
\subfloat{\includegraphics[width = 0.07\textwidth]{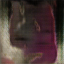}} &
\subfloat{\includegraphics[width = 0.07\textwidth]{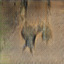}} &
\subfloat{\includegraphics[width = 0.07\textwidth]{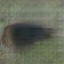}} &
\subfloat{\includegraphics[width = 0.07\textwidth]{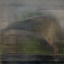}}\\[-1em]
\raisebox{1.25em}{TV} & \subfloat{\includegraphics[width = 0.07\textwidth]{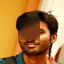}} &
\subfloat{\includegraphics[width = 0.07\textwidth]{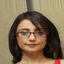}} &
\subfloat{\includegraphics[width = 0.07\textwidth]{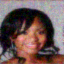}} &
\subfloat{\includegraphics[width = 0.07\textwidth]{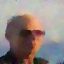}} &
\subfloat{\includegraphics[width = 0.07\textwidth]{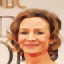}} &
\subfloat{\includegraphics[width = 0.07\textwidth]{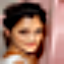}} &
\subfloat{\includegraphics[width = 0.07\textwidth]{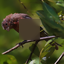}} &
\subfloat{\includegraphics[width = 0.07\textwidth]{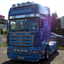}} &
\subfloat{\includegraphics[width = 0.07\textwidth]{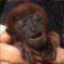}} &
\subfloat{\includegraphics[width = 0.07\textwidth]{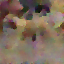}} &
\subfloat{\includegraphics[width = 0.07\textwidth]{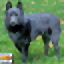}} &
\subfloat{\includegraphics[width = 0.07\textwidth]{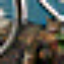}}\\
\end{tabular}
\caption{\small Visual demonstration of inverse problem solutions on the CelebA and STL10 dataset. The problems are identical to the CIFAR-10 case.}\label{fig:celeb_stl_pics}
\end{figure}

A qualitative comparison on sample CIFAR-10 images for the proposed inverse problems is presented in Figure~\ref{fig:cifar_qual}, and similar qualitative comparisons for CelebA and STL10 are presented in Figure~\ref{fig:celeb_stl_pics}.

\end{document}